\documentclass{article}
     \PassOptionsToPackage{sort, numbers, compress}{natbib}
\usepackage[final]{neurips_2023}

\usepackage{amsmath}
\usepackage{amsthm}
\usepackage{amssymb}
\usepackage{neurips_2023}
\usepackage{subfigure}
\usepackage[utf8]{inputenc} 
\usepackage[T1]{fontenc}    
\usepackage{hyperref}       
\usepackage{url}  
\usepackage{booktabs}       
\usepackage{amsfonts}       
\usepackage{nicefrac}       
\usepackage{microtype}      
\usepackage{xcolor}         
\usepackage{graphicx}
\usepackage{wasysym}
\usepackage{wrapfig}
\usepackage{graphicx}
\usepackage{bbm}
\usepackage{bm}
\usepackage{booktabs}
\usepackage{colortbl}
\usepackage[ruled]{algorithm2e}
\usepackage{multirow}
\usepackage{tcolorbox}
\usepackage{float}
\usepackage{color}
\usepackage[misc]{ifsym}
\usepackage{enumerate}

\definecolor{mygray}{gray}{0.9}
\definecolor{mygray1}{gray}{0.95}
\hypersetup{
	colorlinks=true,
	linkcolor=red,
	filecolor=blue,      
	urlcolor=red,
	citecolor=green,
}
\newtheorem{thm}{Theorem}

\newtheorem{proposition}{Proposition}

\newcommand{\etal}{\textit{et al}.}
\newcommand{\ie}{\textit{i}.\textit{e}.}
\newcommand{\eg}{\textit{e}.\textit{g}.}
\newcommand*{\dif}{\mathop{}\!\mathrm{d}}
\newcommand{\y}{\mathbf{y}}
\newcommand{\x}{\mathbf{x}}
\newcommand{\E}{\mathbb{E}}
\title{Optimal Transport-Guided Conditional \\ Score-Based Diffusion Model}

\author{
	Xiang Gu\textsuperscript{1}, Liwei Yang\textsuperscript{1}, Jian Sun (\Letter)\textsuperscript{1,2,3}, Zongben Xu\textsuperscript{1,2,3}\\
	{\textsuperscript{1} School of Mathematics and Statistics, Xi'an Jiaotong University, Xi'an, China}\\
	{\textsuperscript{2} Pazhou Laboratory (Huangpu), Guangzhou, China}\\
	{\textsuperscript{3} Peng Cheng Laboratory, Shenzhen, China}\\
	{\tt\small \{xianggu,yangliwei\}@stu.xjtu.edu.cn}  {\tt\small\{jiansun,zbxu\}@xjtu.edu.cn}
}

\begin{document}

\maketitle

\begin{abstract}
    Conditional score-based diffusion model (SBDM) is for conditional generation of target data with paired data as condition, and has achieved great success in image translation. However, it requires the paired data as condition, and there would be insufficient paired data provided in real-world applications. To tackle the applications with partially paired or even unpaired dataset, we propose a novel Optimal Transport-guided Conditional Score-based diffusion model (OTCS) in this paper. We build the coupling relationship for the unpaired or partially paired dataset based on $L_2$-regularized unsupervised or semi-supervised optimal transport, respectively. Based on the coupling relationship, we develop the objective for training the conditional score-based model for unpaired or partially paired settings, which is based on a reformulation and generalization of the conditional SBDM for paired setting. With the estimated coupling relationship, we effectively train the conditional score-based model by designing  a ``resampling-by-compatibility'' strategy to choose the sampled data with high compatibility as guidance. Extensive experiments on unpaired super-resolution and semi-paired image-to-image translation demonstrated the effectiveness of the proposed OTCS model. From the viewpoint of optimal transport, OTCS provides an approach to transport data across distributions, which is a challenge for OT on large-scale datasets. We theoretically prove that OTCS realizes the data transport in OT with a theoretical bound. Code is available at \url{https://github.com/XJTU-XGU/OTCS}.
\end{abstract}

\section{Introduction}\label{sec:introduction}
Score-based diffusion models (SBDMs)~\cite{song2019generative,ho2020denoising,song2020score,nichol2021improved,bao2022analyticdpm,NEURIPS2021_49ad23d1,karras2022elucidating,Zhao_2023_ICCV} have gained much attention in data generation. SBDMs perturb target data to a Gaussian noise by a diffusion process and learn the reverse process to transform the noise back to the target data. The conditional SBDMs~\cite{dhariwal2021diffusion,ho2020denoising,saharia2022image,whang2022deblurring,saharia2022photorealistic,ho2021classifierfree} that are conditioned on class labels, text, low-resolution images, \textit{etc.}, have shown great success in image generation and translation. The condition data and target data in the conditional SBDMs~\cite{dhariwal2021diffusion,ho2020denoising,saharia2022image,whang2022deblurring,saharia2022photorealistic,ho2021classifierfree} are often paired. That is, we are given a condition for each target sample in training, \eg, in the super-resolution~\cite{saharia2022image,Zhao_2022_CVPR,Zhao_2023_ICCV2}, each high-resolution image (target data) in training is paired with its corresponding low-resolution image (condition).
 However, in real-world applications, there could not be sufficient paired training data, due to the labeling burden. Therefore, it is important and valuable to develop SBDMs for applications with only unpaired or partially paired training data, \eg, unpaired~\cite{zhu2017unpaired} or semi-paired~\cite{mustafa2020transformation} image-to-image translation (I2I). Though there are several SBDM-based approaches~\cite{choi2021ilvr,meng2021sdedit,zhao2022egsde,su2022dual,Yu2023constructing} for unpaired I2I, the score-based models in these approaches are often unconditioned, and the conditions are imposed in inference by cycle consistency~\cite{su2022dual}, designing the initial states~\cite{meng2021sdedit,Yu2023constructing}, or adding a guidance term to the output of the unconditional score-based model~\cite{choi2021ilvr,zhao2022egsde}. It is unclear how to train the conditional score-based model with unpaired training dataset. For the task with a few paired and a large number of unpaired data, \ie, partially paired dataset, there are few SBDMs for tackling this task, to the best of our knowledge.

This paper works on how to train the conditional score-based model with unpaired or partially paired training dataset. We consider the I2I applications in this paper where the condition data and target data are images respectively from different domains.
The main challenges for this task are: 1) the lack of the coupling relationship between condition data and target data hinders the training of the conditional score-based model, and 2) it is unclear how to train the conditional score-based model even with an estimated coupling relationship, because it may not explicitly provide condition-target data pairs as in the setting with paired data. We propose a novel Optimal Transport-guided Conditional Score-based diffusion model (OTCS) to address these challenges. Note that different from the existing OT-related SBDMs that aim to understand~\cite{khrulkov2023understanding} or promote~\cite{li2023dpmot,de2021diffusion} the unconditional score-based models, our approach aims to develop the conditional score-based model for unpaired or partially paired data settings guided by OT.

We tackle the first challenge based on optimal transport (OT). Specifically, for applications with unpaired setting, \eg, unpaired super-resolution, practitioners often attempt to translate the condition data (\eg, low-resolution images) to target domain (\eg, consisting of high-resolution images) while preserving image structures~\cite{daniels2021score},  \textit{etc}. 
We handle this task using unsupervised OT~\cite{villani2009optimal} that transports data points across distributions with the minimum transport cost. The coupling relationship of condition data and target data is modeled in the transport plan of unsupervised OT.  
For applications with partially paired setting, it is reasonable to utilize the paired data to guide building the coupling relationship of unpaired data, because the paired data annotated by humans should have a reliable coupling relationship. The semi-supervised OT~\cite{gu2022keypointguided} is dedicated to leveraging the annotated keypoint pairs to guide the matching of the other data points. So we build the coupling relationship for partially paired dataset using semi-supervised OT by taking the paired data as keypoints.

To tackle the second challenge, we first provide a reformulation of the conditional SBDM for paired setting, in which the coupling relationship of paired data is explicitly considered. Meanwhile, the coupling relationship in this reformulation is closely related to the formulation of the coupling from $L_2$-regularized OT. This enables us to generalize the objective of the conditional SBDM for paired setting to unpaired and partially paired settings based on OT. To train the conditional score-based model using mini-batch data, directly applying the standard training algorithms to our approach can lead to sub-optimal performance of the trained conditional score-based model.
To handle this challenge, we propose a new ``resampling-by-compatibility'' strategy to choose sampled data with high compatibility as guidance in training, which shows effectiveness in experiments.

We conduct extensive experiments on unpaired super-resolution and semi-paired I2I tasks, showing the effectiveness of the proposed OTCS for both applications with unpaired and partially paired settings. 
From the viewpoint of OT, the proposed OTCS offers an approach to transport the data points across distributions, which is known as a challenging problem in OT on large-scale datasets. The data transport in our approach is realized by generating target samples from the optimal conditional transport plan given a source sample, leveraging the capability of SBDMs for data generation.
Theoretically and empirically, we show that OTCS can generate samples from the optimal conditional transport plan of the $L_2$-regularized unsupervised or semi-supervised OTs.

\section{Background}\label{sec:background}
Our method is closely related to OT and conditional SBDMs, which will be introduced below. 

\subsection{Conditional SBDMs with Paired Data}\label{sec:back_sgm}
The conditional SBDMs~\cite{dhariwal2021diffusion,ho2020denoising,saharia2022image,whang2022deblurring,saharia2022photorealistic,NEURIPS2021_49ad23d1,ho2021classifierfree} aim to generate a target sample $\y$ from the distribution $q$ of target training data 
given a condition data $\x$. For the paired setting, each target training sample $\y$ is paired with a condition data $\x_{\rm cond}(\y)$.
The conditional SBDMs with paired dataset can be roughly categorized into two types, respectively under the classifier guidance~\cite{NEURIPS2021_49ad23d1} and classifier-free guidance~\cite{dhariwal2021diffusion,ho2020denoising,saharia2022image,whang2022deblurring,saharia2022photorealistic,ho2021classifierfree}. 
Our approach is mainly related to the second type of methods. These methods use a forward stochastic differential equation (SDE) to add Gaussian noises to the target training data for training the conditional score-based model. The forward SDE is 
$\dif \mathbf{y}_t = f(\mathbf{y}_t,t)\dif t + g(t)\dif \mathbf{w} \mbox{ with }\mathbf{y}_0\sim q$,  and $t\in [0,T]$,
where $\mathbf{w}\in\mathbb{R}^D$ is a standard Wiener process, $f(\cdot,t):\mathbb{R}^D\rightarrow\mathbb{R}^D$ is the drift coefficient, and $g(t)\in\mathbb{R}$ is the diffusion coefficient. 
Let $p_{t|0}$ be the conditional distribution of $\y_t$ given the initial state $\y_0$, and $p_t$ be the marginal distribution of $\y_t$.
We can choose the $f(\mathbf{y},t)$, $g(t)$, and $T$ such that $\mathbf{y}_t$ approaches some analytically tractable prior distribution $p_{\rm prior}(\mathbf{y}_T)$ at time $t=T$, \ie, $p_T(\mathbf{y}_T)\approx p_{\rm{prior}}(\mathbf{y}_T)$. We take $f,g,T$, and $p_{\rm{prior}}$ from two popular SDEs, \ie, VE-SDE and VP-SDE~\cite{song2020score} (please refer to Appendix A for model details). 
The conditional score-based model is trained by denoising score-matching loss:
\begin{equation}\label{eq:denoising_score_matching0}
    \mathcal{J}_{\rm DSM}(\theta) = \mathbb{E}_{t}w_t \mathbb{E}_{\y_0\sim q}\mathbb{E}_{\y_t\sim p_{t|0}(\y_t|\y_0)}
    \left\Vert s_{\theta}(\y_t;\x_{\rm cond}(\y_0),t) - \nabla_{\y_t}\log p_{t|0}(\y_t|\y_0)\right\Vert_2^2,
\end{equation}
where $w_t$ is the weight for time $t$. In this paper, $t$ is uniformly sampled from $[0,T]$, \ie, $t\sim \mathcal{U}([0,T])$. 
With the trained $s_{\hat{\theta}}(\y;\mathbf{x},t)$, given a condition data $\x$, the  target sample $\y_0$ is generated by the reverse SDE as $ \dif \mathbf{y}_t = \left[f(\mathbf{y}_t,t)-g(t)^2s_{\hat{\theta}}(\mathbf{y}_t;\x,t)\right]\dif t + g(t)\dif \bar{\mathbf{w}}$,
where $\bar{\mathbf{w}}$ is a standard Wiener process in the reverse-time direction.
This process starts from a noise sample $\mathbf{y}_T$ and ends at $t=0$.


\subsection{Optimal Transport} \label{sec:back_ot}
\paragraph{Unsupervised OT.} We consider a source distribution $p$ and a target distribution $q$. The unsupervised OT~\cite{kantorovich1942translocation} aims to find the optimal coupling/transport plan $\pi$, \ie, a joint distribution of $p$ and $q$, such that the transport cost is minimized, formulated as the following optimization problem:
\begin{equation}\label{eq:unsupervised_ot}
        \min_{\pi\in\Gamma}\mathbb{E}_{(\mathbf{x},\mathbf{y})\sim\pi}c(\mathbf{x},\mathbf{y}), \mbox{ s.t. } \Gamma=\{\pi:T_{\#}^\mathbf{x}\pi=p, T_{\#}^\mathbf{y}\pi=q\},
\end{equation}
where $c$ is the cost function. 
$T_{\#}^\mathbf{x}\pi$ is the marginal distribution of $\pi$ \textit{w.r.t.} random variable $\mathbf{x}$. $T_{\#}^\mathbf{x}\pi=p$ means $\int \pi(\mathbf{x},\mathbf{y})\dif \mathbf{y}=p(\mathbf{x}), \forall \mathbf{x} \in\mathcal{X}$. Similarly, $T_{\#}^\mathbf{y}\pi=q$ indicates $\int \pi(\mathbf{x},\mathbf{y})\dif \mathbf{x}=q(\mathbf{y}), \forall \mathbf{y} \in\mathcal{Y}$. 
\paragraph{Semi-supervised OT.}The semi-supervised OT is pioneered by~\cite{gu2022keypointguided,Gu2023optimal}.  In semi-supervised OT, a few matched pairs of source and target data points (called ``keypoints'') $\mathcal{K}=\{(\mathbf{x}_k,\mathbf{y}_k)\}_{k=1}^K$ are given, where $K$ is the number of keypoint pairs. The semi-supervised OT aims to leverage the given matched keypoints to guide the correct transport in OT by preserving the relation of each data point to the keypoints. Mathematically, we have
\begin{equation}\label{eq:semi_unsupervised_ot}
\min_{\tilde{\pi}\in\tilde{\Gamma}}\mathbb{E}_{(\mathbf{x},\mathbf{y})\sim m\otimes\tilde{\pi}}g(\mathbf{x},\mathbf{y}), \mbox{ s.t. } \tilde{\Gamma}=\{\tilde{\pi}:T_{\#}^\mathbf{x}(m\otimes\tilde{\pi})=p, T_{\#}^\mathbf{y}(m\otimes\tilde{\pi})=q\},
\end{equation}
where the transport plan $m\otimes\tilde{\pi}$ is $(m\otimes\tilde{\pi})(\mathbf{x},\mathbf{y})=m(\mathbf{x},\mathbf{y})\tilde{\pi}(\mathbf{x},\mathbf{y})$, and $m$ is a binary mask function. Given a pair of keypoints $(\mathbf{x}_{k_0},\mathbf{y}_{k_0})\in \mathcal{K}$, then $m(\mathbf{x}_{k_0},\mathbf{y}_{k_0})=1$, $m(\mathbf{x}_{k_0},\mathbf{y})=0$ for $\mathbf{y}\neq \mathbf{y}_{k_0}$, and $m(\mathbf{x},\mathbf{y}_{k_0})=0$ for $\mathbf{x}\neq \mathbf{x}_{k_0}$. $m(\mathbf{x},\mathbf{y}) = 1$ if $\mathbf{x},\mathbf{y}$ do not coincide with any keypoint. The mask-based modeling of the transport plan ensures that the keypoint pairs are always matched in the derived transport plan. $g$ in Eq.~\eqref{eq:semi_unsupervised_ot} is defined as $g(\mathbf{x},\mathbf{y})=d(R^s_\mathbf{x}, R^t_{\mathbf{y}})$, where $R^s_\mathbf{x}, R^s_{\mathbf{y}}\in (0,1)^K$ model the vector of relation of $\mathbf{x}$, $\mathbf{y}$ to each of the paired keypoints in source and target domain respectively, and $d$ is the Jensen–Shannon divergence. The $k$-th elements of $R^s_\mathbf{x}$ and $R^s_\mathbf{x}$ are respectively defined by 
\begin{equation}\label{eq:relation}
    R^s_{\mathbf{x},k}=\frac{\exp(-c(\mathbf{x},\mathbf{x}_k)/\tau)}{\sum_{l=1}^K\exp(-c(\mathbf{x},\mathbf{x}_{l})/\tau)},\mbox{ } 
    R^t_{\mathbf{y},k}=\frac{\exp(-c(\mathbf{y},\mathbf{y}_k)/\tau)}{\sum_{l=1}^K\exp(-c(\mathbf{y},\mathbf{y}_{l})/\tau)},
\end{equation}
where $\tau$ is set to 0.1. 
Note that, to ensure feasible solutions, the mass of paired keypoints should be equal, \ie, $p(\mathbf{x}_k)=q(\mathbf{y}_k), \forall (\mathbf{x}_k,\mathbf{y}_k)\in\mathcal{K}$. 
Please refer to~\cite{gu2022keypointguided} for more details.

\paragraph{$L_2$-regularized unsupervised and semi-supervised OTs.} As in Eqs.~\eqref{eq:unsupervised_ot} and~\eqref{eq:semi_unsupervised_ot}, both the unsupervised and semi-supervised OTs are linear programs that are computationally expensive to solve for larger sizes of training datasets. Researchers then present the $L_2$-regularized versions of unsupervised and semi-supervised OTs that can be solved by training networks~\cite{seguy2017large,gu2023keypointguided}. 
The $L_2$-regularized unsupervised and semi-supervised OTs are respectively given by
\begin{equation}\label{eq:l2_regularized}
    \min_{\pi\in\Gamma}{\mathbb{E}}_{(\mathbf{x},\mathbf{y})\sim\pi}c(\mathbf{x},\mathbf{y})+\epsilon\chi^2(\pi\Vert p\times q) \mbox{ and }
    \min_{\tilde{\pi}\in\tilde{\Gamma}}{\mathbb{E}}_{(\mathbf{x},\mathbf{y})\sim m\otimes\tilde{\pi}}g(\mathbf{x},\mathbf{y})+\epsilon\chi^2(m\otimes\tilde{\pi}\Vert p\times q),
\end{equation}
where $\chi^2(\pi\Vert p\times q) = \int \frac{\pi(\mathbf{x},\mathbf{y})^2}{p(\mathbf{x})q(\mathbf{y})}\dif \mathbf{x}\dif \mathbf{y}$, $\epsilon$ is regularization factor. The duality of the $L_2$-regularized unsupervised and semi-supervised OTs can be unified in the following formulation~\cite{seguy2017large,gu2023keypointguided}:
\begin{equation}\label{eq:duality}
    \max_{u,v}\mathcal{F}_{\rm OT}(u,v) = \mathbb{E}_{\mathbf{x}\sim p}u(\mathbf{x}) + \mathbb{E}_{\mathbf{y}\sim q}v(\mathbf{y}) - \frac{1}{4\epsilon}\mathbb{E}_{\mathbf{x}\sim p,\mathbf{y}\sim q}I(\mathbf{x},\mathbf{y})\left[\left(u(\mathbf{x})+v(\mathbf{y})-\xi(\mathbf{x},\mathbf{y})\right)_+\right]^2,
\end{equation}
where $a_+=\max(a,0)$. For unsupervised OT, $I(\mathbf{x},\mathbf{y})=1$ and $\xi(\mathbf{x},\mathbf{y})=c(\mathbf{x},\mathbf{y})$. For semi-supervised OT, $I(\mathbf{x},\mathbf{y})=m(\mathbf{x},\mathbf{y})$ and $\xi(\mathbf{x},\mathbf{y})=g(\mathbf{x},\mathbf{y})$. In~\cite{seguy2017large,gu2023keypointguided}, $u,v$ are represented by neural networks $u_{\omega},v_{\omega}$ with parameters $\omega$ that are trained by mini-batch-based stochastic optimization algorithms using the loss function in Eq.~\eqref{eq:duality}. The pseudo-code for training $u_{\omega},v_{\omega}$ is given in Appendix A.  Using the parameters ${\hat{\omega}}$ after training, the estimate of optimal transport plan is 
\begin{equation}\label{eq:estimate_plan}
    \hat{\pi}(\mathbf{x},\mathbf{y})=H(\mathbf{x},\mathbf{y})p(\mathbf{x})q(\mathbf{y}), \mbox{ where } H(\mathbf{x},\mathbf{y}) = \frac{1}{2\epsilon}I(\mathbf{x},\mathbf{y})\left(u_{\hat{\omega}}(\mathbf{x})+v_{\hat{\omega}}(\mathbf{y})-\xi(\mathbf{x},\mathbf{y})\right)_+.
\end{equation}
$H$ is called compatibility function. 
Note that $\hat{\pi}$ and $H$ depend on $c$ and $\epsilon$, which will be specified in experimental details in Appendix C. 

\begin{figure}[t]
    \centering
    \includegraphics[width=1.0\columnwidth]{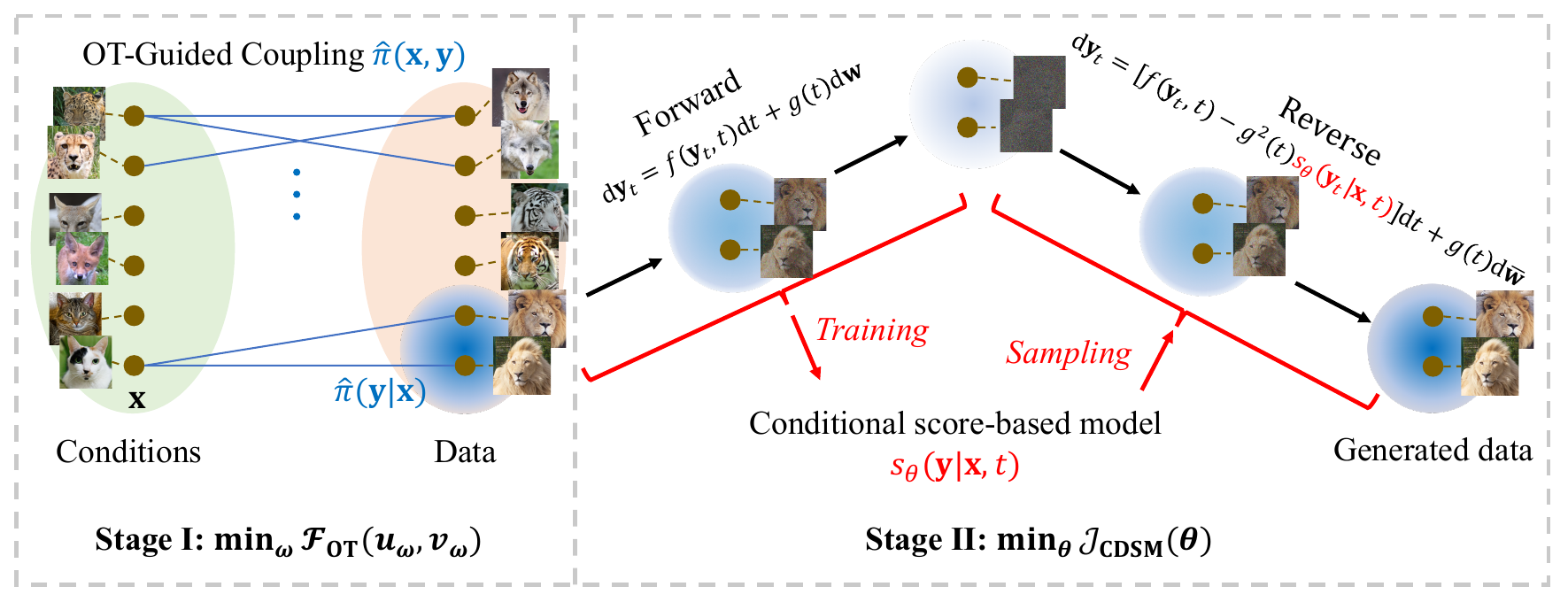} 
    \caption{Illustation of optimal transport-guided conditional score-based diffusion model. We build the coupling $\hat{\pi}(\x,\y)$ of condition data (\eg, source images in I2I) $\x$ and target data $\y$ guided by OT. Based on the coupling, we train the conditional score-based model $s_{\theta}(\y;\x,t)$ by OT-guided conditional denoising score-matching that uses the forward SDE to diffuse target data to noise. With $s_{\theta}(\y;\x,t)$, we generate data given the condition $\x$ using the reverse SDE in inference.}
    \label{fig:big_figure}
\end{figure}

\section{OT-Guided Conditional SBDM}\label{sec:method}
We aim to develop conditional SBDM for applications with unpaired or partially paired setting. 
The unpaired setting means that there is no paired condition data and target data in training. For example, in unpaired image-to-image translation (I2I), the source and target data in the training set are all unpaired. For the partially paired setting, along with the unpaired set of condition data (e.g., source data in I2I) and target data, we are also given a few paired condition data and target data. 

We propose an Optimal Transport-guided Conditional SBDM, dubbed OTCS, for conditional diffusion in both unpaired and partially paired settings. The basic idea is illustrated in Fig.~\ref{fig:big_figure}.
Since the condition data and target data are not required to be paired in the training dataset, we build the coupling relationship of condition and target data using OT~\cite{villani2009optimal,gu2022keypointguided}. 
Based on the estimated coupling, we propose the OT-guided conditional denoising score matching to train the conditional score-based model in the unpaired or partially paired setting.
With the trained conditional score-based model, we generate a sample by the reverse SDE given the condition. We next elaborate on the motivations, OT-guided conditional denoising score matching, training, and inference of our approach.

\subsection{Motivations for OT-Guided Conditional SBDM}\label{sec:motivations}
We provide a reformulation for the conditional score-based model with paired training dataset discussed in Sect.~\ref{sec:back_sgm}, and this formulation motivates us to extend the conditional SBDM to unpaired and partially paired settings in Sect.~\ref{sec:train_score_model}. Let $q$ and $p$ respectively denote the distributions of target data and condition data. 
In the paired setting, we denote the condition data as $\x_{\rm cond}(\y)$ for a target data $\y$, and $p$ is the measure by push-forwarding $q$ using $\x_{\rm cond}$ , \ie, $p(\x) =\sum_{\{\y:\x_{\rm cond}(\y) = \x\}}q(\y)$
over the paired training dataset.
\begin{proposition}\label{lem:lemma1}
Let $\mathcal{C}(\x,\y)=\frac{1}{p(\x)}\delta(\x-\x_{\rm cond}(\y))$ where $\delta$ is the Dirac delta function, then $\mathcal{J}_{\rm DSM}(\theta)$ in Eq.~\eqref{eq:denoising_score_matching0} can be reformulated as
\begin{equation}\label{eq:reformulation_paired_data}
    \mathcal{J}_{\rm DSM}(\theta)=\E_tw_t\E_{\x\sim p}\E_{\y\sim q}\mathcal{C}(\x,\y) \E_{\y_t\sim p_{t|0}(\y_t|\y)}
    \left\Vert s_{\theta}(\y_t;\x,t) - \nabla_{\y_t}\log p_{t|0}(\y_t|\y)\right\Vert_2^2.
\end{equation}
Furthermore, $\gamma(\x,\y)=\mathcal{C}(\x,\y)p(\x)q(\y)$ is a joint distribution for marginal distributions $p$ and $q$. 
\end{proposition}
The proof is given in Appendix B.
From Proposition~\ref{lem:lemma1}, we have the following observations. First, the coupling relationship of condition data and target data is explicitly modeled in $\mathcal{C}(\x,\y)$. Second, the joint distribution $\gamma$ exhibits a similar formulation to the transport plan $\hat{\pi}$ in Eq.~\eqref{eq:estimate_plan}. The definition of $\mathcal{C}(\x,\y)$ in Proposition~\ref{lem:lemma1} is for paired $\x, \y$. While for the unpaired or partially paired setting, the definition of $\mathcal{C}(\x,\y)$ is not obvious due to the lack of paired relationship between $\x, \y$. We therefore consider modeling the joint distribution of condition data ($\x$) and target data ($\y$) by $L_2$-regularized OT (see Sect.~\ref{sec:back_ot}) for the unpaired and partially paired settings, in which the coupling relationship of condition data and target data is built in the compatibility function $H(\x,\y)$. 

\subsection{OT-Guided Conditional Denoising Score Matching} \label{sec:train_score_model}
With the motivations discussed in Sects.~\ref{sec:introduction} and~\ref{sec:motivations}, we model the coupling relationship between condition data and target data for unpaired and partially paired settings using $L_2$-regularized unsupervised and semi-supervised OTs, respectively. Specifically, the $L_2$-regularized unsupervised and semi-supervised OTs are applied to the distributions $p,q$, and the coupling relationship of the condition data $\x$ and target data $\y$ is built by the compatibility function $H(\x,\y)$. We then extend the formulation for  paired setting in Eq.~\eqref{eq:reformulation_paired_data} by replacing $\mathcal{C}$ with $H$ to develop the training objective for unpaired and partially paired settings, which is given by 
\begin{equation}\label{eq:conditional_denoising_score_matching1} 
    \mathcal{J}_{\rm CDSM}(\theta)  = 
    \mathbb{E}_{t}w_t{\mathbb{E}}_{\x\sim p}{\mathbb{E}}_{\y\sim q}H(\x,\y) {\mathbb{E}}_{\y_t\sim p_{t|0}(\y_t|\y)}\left\Vert s_{\theta}(\mathbf{y}_t;\x,t) - \nabla_{\y_t}\log p_{t|0}(\y_t|\y)\right\Vert_2^2.
\end{equation} Equation~\eqref{eq:conditional_denoising_score_matching1} is dubbed ``OT-guided conditional denoising score matching''. In Eq.~\eqref{eq:conditional_denoising_score_matching1}, $H$ is a ``soft'' coupling relationship of condition data and target data, because there may exist multiple $\x$ satisfying $H(\x,\y)>0$ for each $\y$. While Eq.~\eqref{eq:reformulation_paired_data} assumes ``hard'' coupling relationship, \ie, there is only one condition data $\x$ for each $\y$ satisfying $\mathcal{C}(\x,\y)>0$. We minimize $\mathcal{J}_{\rm CDSM}(\theta)$ to train the conditional score-based model $s_{\theta}(\mathbf{y}_t;\x,t)$. We will theoretically analyze that our formulation in Eq.~\eqref{eq:conditional_denoising_score_matching1} is still a diffusion model in Sect.~\ref{sec:analysis}, and empirically compare the ``soft'' and ``hard'' coupling relationship in Appendix~D.

\subsection{Training the Conditional Score-based Model}\label{sec:training}
To implement $\mathcal{J}_{\rm CDSM}(\theta)$ in Eq.~\eqref{eq:conditional_denoising_score_matching1} using training samples to optimize $\theta$, we can sample mini-batch data $\bm{X}$ and $\bm{Y}$ from $p$ and $q$ respectively, and then compute $H(\x,\y)$ and 
$ \mathcal{J}_{\x,\y}=\E_tw_t {\mathbb{E}}_{\y_t\sim p_{t|0}(\y_t|\y)}$ $\Vert s_{\theta}(\mathbf{y}_t;\x,t) - \nabla_{\y_t}\log p_{t|0}(\y_t|\y)\Vert_2^2$
over the pairs of $(\x,\y)$ in $\bm{X}$ and $\bm{Y}$. However, such a strategy is sub-optimal. This is because given a mini-batch of samples $\bm{X}$ and $\bm{Y}$, for each source sample $\x$, there may not exist target sample $\y$ in the mini-batch with a higher value of $H(\x,\y)$ that matches condition data $\x$. Therefore, few or even no samples in a mini-batch contribute to the loss function in Eq.~\eqref{eq:conditional_denoising_score_matching1}, leading to a large bias of the computed loss and instability of the training. To tackle this challenge, we propose a ``resampling-by-compatibility'' strategy to compute the loss in Eq.~\eqref{eq:conditional_denoising_score_matching1}.

\paragraph{Resampling-by-compatibility.} To implement the loss in Eq.~\eqref{eq:conditional_denoising_score_matching1}, we perform the following steps:
\begin{itemize}
    \item Sample $\x$ from $p$ and sample $\bm{Y}_{\x}=\{\y^l\}_{l=1}^L$ from $q$;
    \item Resample a $\y$ from $\bm{Y}_{\x}$ with the probability proportional to $H(\x,\y^l)$;
    \item Compute the training loss $\mathcal{J}_{\x,\y}$ in the above paragraph on the sampled pair $(\x,\y)$.
\end{itemize}
In implementation, $u_{\omega}$ and $v_{\omega}$ (introduced above Eq.~\eqref{eq:estimate_plan} in Sect.~\ref{sec:back_ot}) are often lightweight neural networks, so $H(\x,\y^l)$ can be computed fast as in Eq.~\eqref{eq:estimate_plan}. In the applications where we are given training datasets of samples, for each $\x$ in the set of condition data, we choose all the samples $\y$ in the set of target data satisfying $H(\x,\y)>0$ to construct $\bm{Y}_{\x}$,
and meanwhile store the corresponding values of $H(\x,\y)$. This is done before training $s_{\theta}$. During training, we directly choose $\y$ from $\bm{Y}_{\x}$ based on the stored values of $H$, which speeds up the training process. 
Please refer to Appendix A for the rationality of the resampling-by-compatibility.

\paragraph{Algorithm.} 
Before training $s_{\theta}$, we train $u_{\omega},v_{\omega}$ using the duality of $L_2$-regularized unsupervised and semi-supervised OTs in Eq.~\eqref{eq:duality} for unpaired and partially paired settings, respectively. 
During training $s_{\theta}$, in each iteration, we sample a mini-batch of condition data $\bm{X} = \{\x_b\}_{b=1}^B$ from $p$ with batch size $B$. For $\x_b$, we sequentially obtain a sample $\y$ using our resampling-by-compatibility strategy, uniformly sample a $t$ in $[0,T]$, and generate a noisy data $\y_t$ from $p_{t|0}(\y_t|\y)$.
We then compute the value of $\mathcal{J}_{\x_b,\y}$ (defined in the first paragraph of this section) on $t$ and $\y_t$, which is averaged over all $b$ as the final training loss.
The parameters $\theta$ of $s_{\theta}$ are then updated by stochastic optimization algorithms, \eg, Adam. The pseudo-code of the training algorithm is given in Appendix~A.


\subsection{Sample Generation}\label{sec:generate_samples}

We denote the 
trained conditional score-based model as $s_{{\hat{\theta}}}(\y;\x,t)$ where $\hat{\theta}$ is the value of $\theta$ after training. Given the condition data $\x$, we generate target samples by the following SDE:
\begin{equation}
\begin{split}\label{eq:reverse_sde_conditional}
    \dif \y_t = \left[f(\y_t,t)-g(t)^2 s_{{\hat{\theta}}}(\y_t;\x,t)\right]\dif t + g(t) \dif\bar{\mathbf{w}}, 
\end{split}
\end{equation}
with the initial state $\y_T\sim p_{\rm prior}$.
The numerical SDE solvers, \eg, Euler-Maruyama method~\cite{platen1999introduction}, DDIM~\cite{song2020denoising}, and DPM-Solver~\cite{lu2022dpmsolver}, are then applied to solve the above reverse SDE.

\subsection{Analysis}\label{sec:analysis}
In this section, we analyze that by Eq.~\eqref{eq:reverse_sde_conditional}, our approach approximately generates samples from the conditional transport plan $\hat{\pi}(\y|\x)$, where $\hat{\pi}(\y|\x)=H(\x,\y)q(\y)$ is based on $\hat{\pi}(\x,\y)$ in Eq.~\eqref{eq:estimate_plan}.

\begin{thm}\label{thm:thm1}
        For $\x\sim p$, we define the forward SDE $\dif \y_t = f(\y_t,t)\dif t + g(t)\dif\mathbf{w}$ with $\y_0\sim\hat{\pi}(\cdot|\x)$ and $t\in [0,T]$, where $f,g,T$ are given in Sect.~\ref{sec:back_sgm}. Let $p_{t}(\y_t|\x)$ be the corresponding distribution of $\y_t$ and
        $\mathcal{J}_{\rm CSM}(\theta) =  \mathbb{E}_{t}w_t\mathbb{E}_{\x\sim p}\mathbb{E}_{\y_t\sim p_t(\y_t|\x)} \left\Vert s_{\theta}(\mathbf{y}_t;\x,t) - \nabla_{\y_t}\log p_t(\y_t|\x)\right\Vert_2^2$,
    then
     we  have  $\nabla_{\theta}\mathcal{J}_{\rm CDSM}(\theta)=\nabla_{\theta}\mathcal{J}_{\rm CSM}(\theta)$.
\end{thm}
We give the proof in Appendix B.
Theorem~\ref{thm:thm1} indicates that the trained $s_{\theta}(\y_t;\x,t)$ using Eq.~\eqref{eq:conditional_denoising_score_matching1} approximates $\nabla_{\y_t}\log p_t(\y_t|\x)$. 
Based on Theorem~\ref{thm:thm1}, we can interpret our approach as follows. Given a condition data $\x$, we sample target data $\y_0$ from the conditional transport plan $\hat{\pi}(\y_0|\x)$, produce $\y_t$ by the forward SDE, and train $s_{\theta}(\y_t;\x,t)$ to approximate $\nabla_{\y_t}\log p_t(\y_t|\x)$, as illustrated in Fig.~\ref{fig:big_figure}. This implies that Eq.~\eqref{eq:reverse_sde_conditional} approximates the reverse SDE $\dif \y_t = [f(\y_t,t)$ $-g(t)^2 \nabla_{\y_t} \log p_t(\y_t|\x)]\dif t + g(t) \dif \bar{\mathbf{w}}$, by which the generated samples at time $t=0$ are from $p_0(\y|\x)=\hat{\pi}(\y|\x)$ given the initial state $\y_T\sim p_T(\y_T|\x)$.

\section{OTCS Realizes Data Transport for Optimal Transport}\label{sec:OTCS_realize_OT}
As discussed in Sect.~\ref{sec:method}, OTCS is proposed to learn the conditional SBDM guided by OT. In this section, we will show that, from the viewpoint of OT, OTCS offers a diffusion-based approach to transport data for OT. Given a source distribution $p(\x)$ and a target distribution $q(\y)$, the derived coupling $\pi(\x,\y)$ models the joint probability density function rather than the transported sample of $\x$. How to transport the source data points to the target domain is known to be a challenging problem for large-scale OT~\cite{genevay2016stochastic,seguy2017large}. Based on $\pi$,  Seguy \etal~\cite{seguy2017large} transport $\x$ to the barycenter of  $\pi(\cdot|\x)$ which is a blurred sample.  Daniels \etal~\cite{daniels2021score} transport $\x$ to a target sample generated from $\pi(\cdot|\x)$. In line with~\cite{daniels2021score},   
we next theoretically show that our proposed OTCS can generate samples from $\pi(\cdot|\x)$. The comparison of OTCS with~\cite{daniels2021score} will be given in the last paragraph of this section.  

We next study the upper bound of the distance between the distribution (denoted as $p^{\rm sde}(\y|\x)$) of generated samples by OTCS and the optimal conditional transport plan $\pi(\y|\x)$.
For convenience, we investigate the upper bound of the expected Wasserstein distance $\E_{\x\sim p}W_2\left(p^{\rm sde}(\cdot|\x),\pi(\cdot|\x)\right)$.
We denote the Lagrange function for the $L_2$-regularized unsupervised or semi-supervised OTs in Eq.~\eqref{eq:duality} as $\mathcal{L}(\pi,u,v)$ with dual variables $u,v$ as follows:
\begin{equation}
\begin{split}
        \mathcal{L} (\pi,u,v) = &\int \left(\xi(\x,\y)\pi(\x,\y) + \epsilon \frac{\pi(\x,\y)^2}{p(\x)q(\y)}\right)\dif\x\dif\y \\
        + &\int u(\x)\left(\int \pi(\x,\y)\dif\y- p(\x)\right)\dif\x + \int v(\y)\left(\int\pi(\x,\y)\dif\x-q(\y)\right)\dif\y.
\end{split}
\end{equation}
For semi-supervised OT, $\pi$ is further constrained by $\pi=m\otimes\tilde{\pi}$. We follow~\cite{daniels2021score} to assume that $\mathcal{L} (\pi,u,v)$ is $\kappa$-strongly convex in $L_1$-norm \textit{w.r.t.} $\pi$, and take the assumptions (see Appendix B) in~\cite{kwon2022scorebased} that investigates the bound for unconditional SBDMs.

\begin{thm}\label{thm:bound}
Suppose the above assumption and the assumptions in Appendix B hold, and $w_t=g(t)^2$, then we have 
\begin{equation}\label{eq:bound}
\begin{split}
    \E_{\x\sim p}W_2\left(p^{\rm sde}(\cdot|\x),\pi(\cdot|\x)\right)
 \leq & C_1 \big\Vert \nabla_{\hat{\pi}}\mathcal{L} (\hat{\pi}, u_{\hat{\omega}}, v_{\hat{\omega}})\big\Vert_1+ \sqrt{C_2 \mathcal{J}_{\rm CSM}(\hat{\theta})} \\
  +& C_3\E_{\x\sim p}W_2\left(p_T(\cdot|\x),p_{\rm prior}\right),
\end{split}
\end{equation}
where $C_1,C_2$, and $C_3$ are constants to $\hat{\omega}$ and $\hat{\theta}$ given in Appendix B.
\end{thm}

The proof is provided in Appendix B.
In OTCS, we use $u_{{\omega}}$ and $v_{{\omega}}$ to respectively parameterize $u$ and $v$ as discussed in Sect.~\ref{sec:back_ot}. The trained $u_{\hat{\omega}}$ and $v_{\hat{\omega}}$ are the minimizer of the dual problem in Eq.~\eqref{eq:duality} that are near to the saddle point of $\mathcal{L}(\pi,u,v)$. This implies that the gradient norm $\left\Vert \nabla_{\hat{\pi}}\mathcal{L} (\hat{\pi}, u_{\hat{\omega}}, v_{\hat{\omega}})\right\Vert_1$ of the Lagrange function \textit{w.r.t.} the corresponding primal variable $\hat{\pi}$ is minimized in our approach. 
The conditional score-based model $s_{\theta}$ is trained to minimize $\mathcal{J}_{\rm CSM}(\theta)$ according to Theorem~\ref{thm:thm1}. So the loss $\mathcal{J}_{\rm CSM}(\hat{\theta})$ of trained $s_{\hat{\theta}}$ is minimized.
We choose the forward SDE such that $p_T(\cdot|\x)$ is close to $p_{\rm prior}$, which minimizes $\E_{\x\sim p}W_2\left(p_T(\cdot|\x),p_{\rm prior}\right)$. Therefore, OTCS minimizes $\E_{\x\sim p}W_2\left(p^{\rm sde}(\cdot|\x),\pi(\cdot|\x)\right)$, indicating that OTCS can approximately generate samples from $\pi(\cdot|\x)$.

\begin{figure}[t]
    \centering
    \subfigure[$\nabla_\mathbf{y}\log H(-4,\mathbf{y})$]{\includegraphics[width=.24\columnwidth]{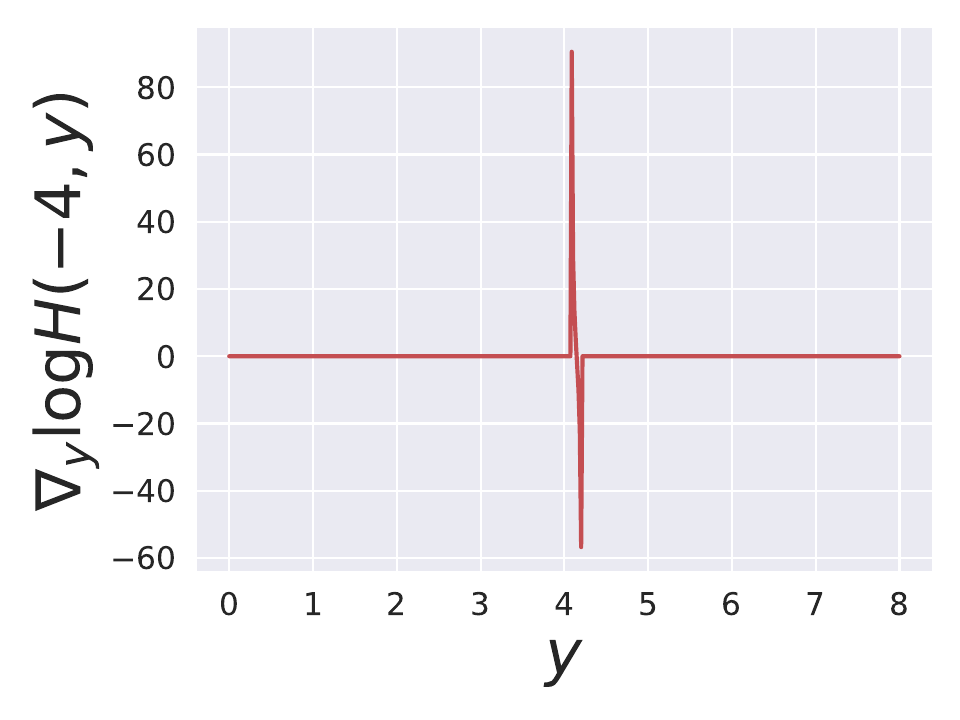}\label{fig:grad_H}}
    \subfigure[SCONES]{\includegraphics[width=0.24\columnwidth]{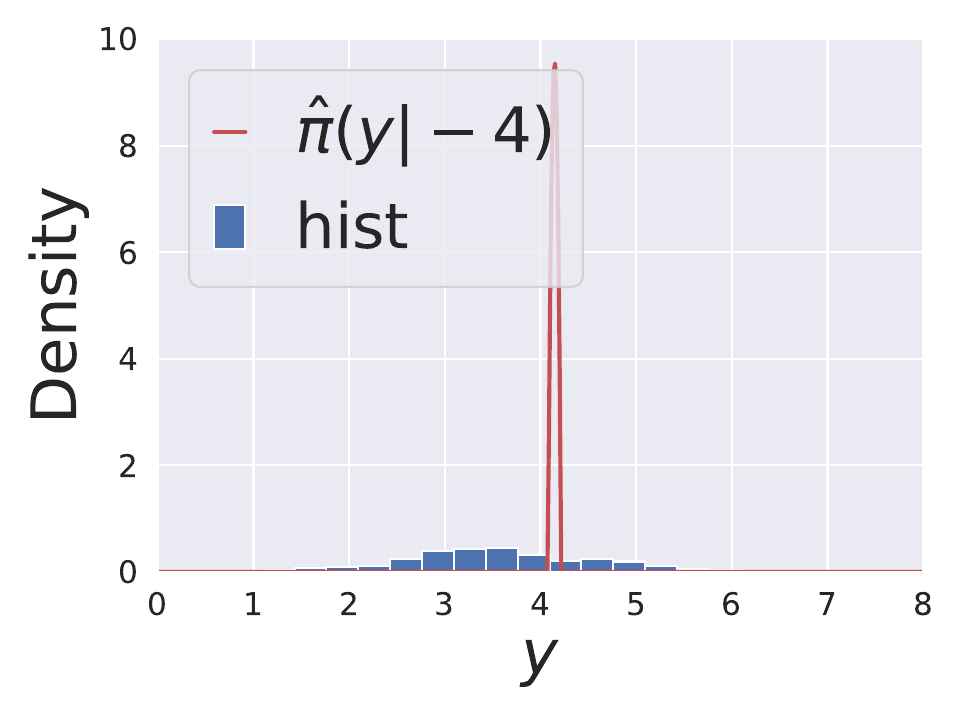}\label{fig:1d_scones}}
    \subfigure[OTCS (ours)]{\includegraphics[width=0.24\columnwidth]{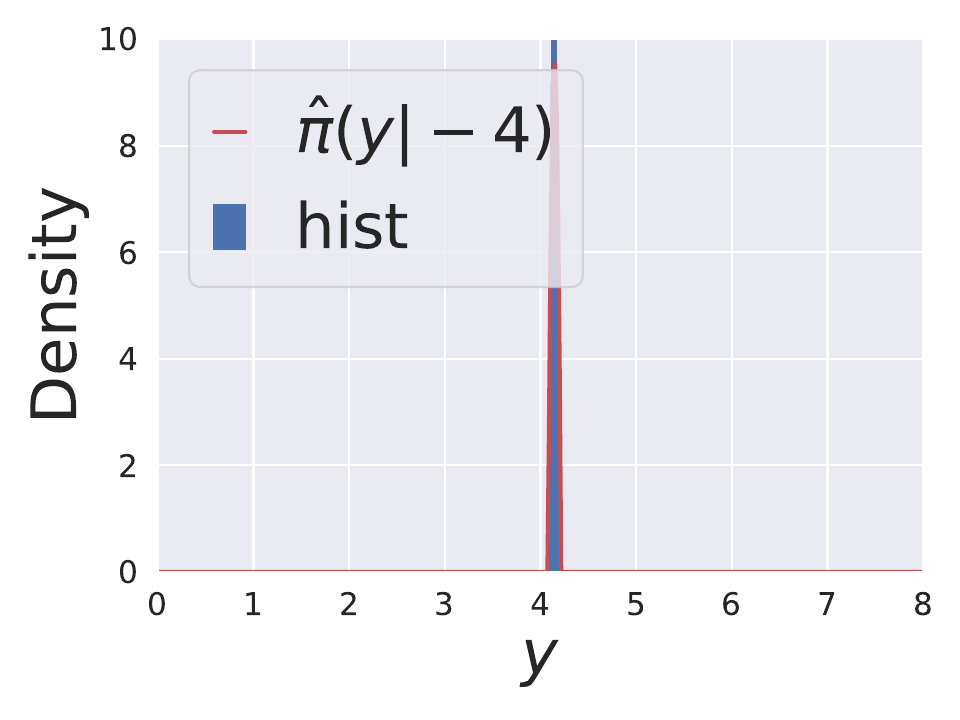}\label{fig:1d_otcs}}
    \subfigure[Distance]{\includegraphics[width=0.24\columnwidth]{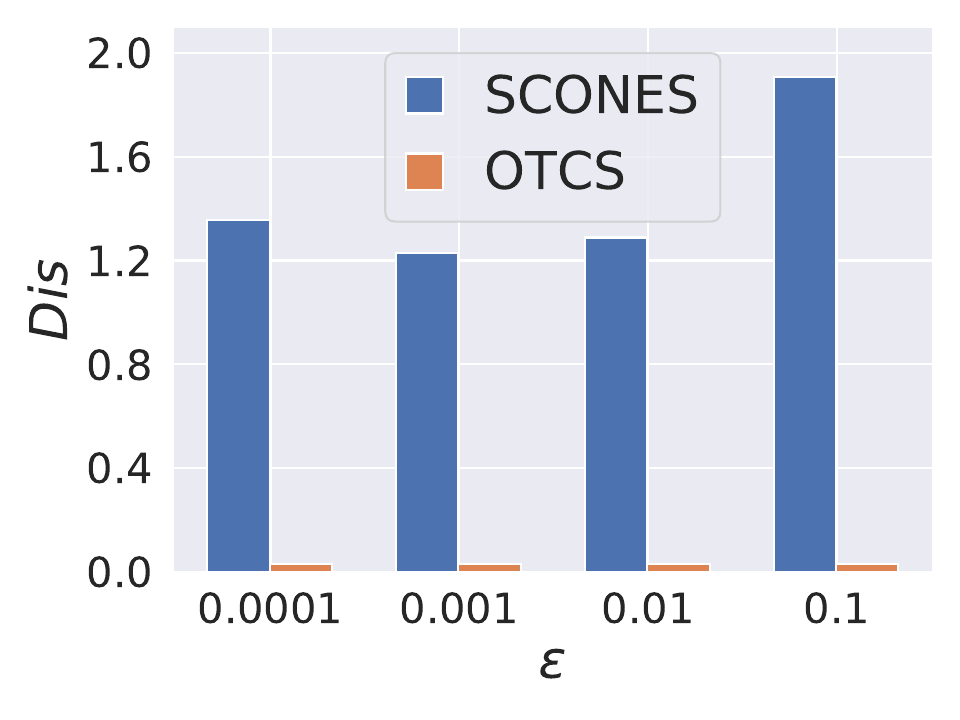}\label{fig:1d_dist}}
    \vspace{-0.2cm}
    \caption{1-D example of $L_2$-regularized unsupervised OT between $p(\mathbf{x})=\mathcal{N}(\mathbf{x}|-4,1)$ and $q(\mathbf{y})=\mathcal{N}(\mathbf{y}|4,1)$. (a) The gradient $\nabla_\mathbf{y}\log H(-4,\mathbf{y})$. (b-c) The conditional transport plan $\hat{\pi}(\mathbf{y}|-4)$ and the histograms of generated samples by (b) SCONES and (c) OTCS with $\epsilon=0.0001$. (d) The distance $Dis=\E_{\x\sim p}W_2\left(p^{\rm sde}(\cdot|\x),\pi(\cdot|\x)\right)$ for different $\epsilon$ computed using samples in which we approximate $p^{\rm sde}(\cdot|\x)$ and $\pi(\cdot|\x)$ by Gaussian distributions with corresponding mean and variance.
    }
    \label{fig:1d_example}
    \vspace{-0.4cm}
\end{figure}

\paragraph{Comparison with related large-scale OT methods.} Recent unsupervised OT methods~\cite{makkuva2020optimal,rout2022generative,korotin2023kernel,daniels2021score} often parameterize the transport map by a neural network learned by adversarial training based on the dual formulation~\cite{makkuva2020optimal,rout2022generative,korotin2023kernel}, or generate samples from the conditional transport plan~\cite{daniels2021score}. Our method is mostly related to SCONES~\cite{daniels2021score} that leverages the unconditional SBDM for generating samples from the estimated conditional transport plan $\hat{\pi}(\y|\x)$. 
Motivated by the expression of $\hat{\pi}(\y|\x)$ in Sect.~\ref{sec:analysis}, given source sample $\x$, SCONES generates target sample $\y$ by the reverse SDE $\dif \mathbf{y}_t = [f(\mathbf{y}_t,t)-g(t)^2 \left(\nabla_{\mathbf{y}_t}\log H(\mathbf{x},\mathbf{y}_t) + s_{\theta}(\mathbf{y}_t;t)\right)]\dif t + g(t)\dif \bar{\mathbf{w}}$, where $s_{\theta}(\mathbf{y}_t;t)$ is the unconditional score-based model. 
The compatibility function $H$ is trained on clean data. While in SCONES, $\nabla_{\mathbf{y}_t}\log H(\mathbf{x},\mathbf{y}_t)$ is computed on noisy data $\mathbf{y}_t$ for $t>0$. OTCS computes $H$ on clean data as in Eq.~\eqref{eq:conditional_denoising_score_matching1}.
We provide a 1-D example in Fig.~\ref{fig:1d_example} to evaluate SCONES and OTCS.
From Figs.~\ref{fig:1d_example}{\color{red}(b-c)}, we can see that the sample histogram, generated by OTCS given ``-4'' as condition, better fits $\hat{\pi}(\cdot|-4)$. In Fig.~\ref{fig:1d_dist}, OTCS achieves a lower expected Wasserstein distance $\E_{\x\sim p}W_2\left(p^{\rm sde}(\cdot|\x),\hat{\pi}(\cdot|\x)\right)$. In SCONES, the gradient of $H$ on noisy data could be inaccurate or zero (as shown in Fig.~\ref{fig:grad_H}), which may fail to guide the noisy data $\y_t$ to move towards the desired locations as $t\rightarrow 0$. Our
OTCS utilizes $H$ to guide the training of the conditional score-based model, without requiring the gradient of $H$ in inference. This may account for the better performance of OTCS.


\section{Experiments}\label{sec:experiments}
We evaluate OTCS on unpaired super-resolution and semi-paired I2I.
 Due to space limits, additional experimental details and results are given in Appendix~C and D, respectively. 

\subsection{Unpaired Super-Resolution on CelebA Faces}\label{sec:super_resolution}
We consider the unpaired super-resolution for 64$\times$64 aligned faces on CelebA dataset~\cite{liu2015deep}. Following~\cite{daniels2021score}, we split the dataset into 3 disjointed subsets: A1 (90K), B1 (90K), and C1 (30K). For images in each subset, we do 
2$\times$ bilinear downsampling to obtain the low-resolution images, followed by 2$\times$ bilinear upsampling to generate datasets of
A0, B0, and C0 correspondingly. 
We train the model on A0 and B1 respectively as the source and target datasets, and test on C0 for generating high-resolution images. 
To apply OTCS to this task, we take B1 as target data and A0 as unpaired source data. We use the $L_2$-regularized unsupervised OT to estimate the coupling, where $c$ is set to the mean squared $L_2$-distance.  In testing, given a degenerated image $\x$ in C0 as condition, we follow~\cite{zhao2022egsde,meng2021sdedit} to sample noisy image $\y_M$ from $p_{M|0}(\y|\x)$ as initial state and perform the reverse SDE to generate the high-resolution image. $M=0.2$ in experiments.
Our approach is compared with recent adversarial-training-based unsupervised OT methods~\cite{jacob*2019wgan,makkuva2020optimal,rout2022generative,korotin2023neural,korotin2023kernel}, diffusion-based unpaired I2I methods~\cite{zhao2022egsde,su2023dual}, flow-based I2I method~\cite{liu2023flow}, and SCONES~\cite{daniels2021score} (discussed in Sect.~\ref{sec:OTCS_realize_OT}).
We use the FID score~\cite{heusel2017gans} to measure the quality of translated images, and the SSIM metric to measure the structural similarity of each translated image to its ground-truth high-resolution image in C1. 

\begin{figure}[t]
\vspace{-0.1cm}
	\centering
        \includegraphics[width=0.99\columnwidth]{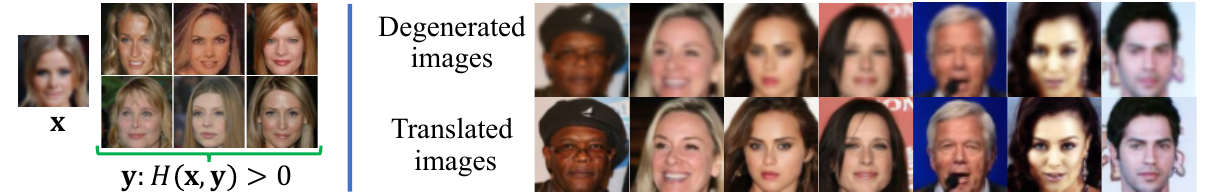}
	\caption{Left: The guided high-resolution images (\ie, $\y:H(\x,\y)>0$) sampled based on OT for low-resolution image $\x$ in training. Right: results of OTCS on CelebA (64$\times$64) in unpaired setting.}
    \label{fig:qualitative_celeba}
    \vspace{-0.4cm}
\end{figure}

In Tab.~\ref{tab:quantitative_results}, OTCS achieves the lowest FID and the highest SSIM on CelebA dataset among the compared methods.
We can observe that in general, the adversarial-training-based OT methods~\cite{jacob*2019wgan,makkuva2020optimal,rout2022generative,korotin2023kernel,korotin2023neural} achieve better SSIM than the diffusion-based methods~\cite{zhao2022egsde,liu2023flow,su2023dual,daniels2021score}. This could be because the OT guidance is imposed in training by these OT methods. By contrast, the diffusion-based methods~\cite{zhao2022egsde,liu2023flow,su2023dual,daniels2021score} generally achieve better FID than OT methods~\cite{jacob*2019wgan,makkuva2020optimal,rout2022generative,korotin2023kernel,korotin2023neural}, which may be attributed to the capability of diffusion models for generating high-quality images. Our OTCS imposes the OT guidance in the training of diffusion models, integrating both advantages of OT and diffusion models.
In Fig.~\ref{fig:qualitative_celeba}, we show the guided high-resolution images sampled based on OT (left) and translated images by OTCS (right).
We also report the FID (0.78) and SSIM (0.9635) of the Oracle that uses true high-resolution images in A1 as paired data for training. We can see that OTCS approaches the Oracle.


\begin{table}[t]
	\centering
	\caption{Quantitative results for unpaired super-resolution on Celeba and semi-paired I2I on Animal images and Digits. The best and second best are respectively bolded and underlined.
	}
	\setlength{\tabcolsep}{5.5pt}

\begin{tabular}{lc|cc|cc|cc}
		\toprule
		\multirow{2}*{Method} &\multirow{2}*{Method Type}& \multicolumn{2}{c|}{Celeba}& \multicolumn{2}{c|}{Animal images}&\multicolumn{2}{c}{Digits}\\
           ~&~&FID $\downarrow$ &SSIM $\uparrow$ &FID $\downarrow$& Acc (\%) $\uparrow$&FID $\downarrow$ & Acc (\%)  $\uparrow$\\
           \midrule
           W2GAN~\cite{jacob*2019wgan} &OT& 48.83&0.7169&118.45&33.56&\ \ 97.06&29.13\\
           OT-ICNN~\cite{makkuva2020optimal} &OT&33.26&0.8904&148.29&38.44&\ \ 50.33&10.84\\
           OTM~\cite{rout2022generative} &OT& 22.93&0.8302&\ \ 69.27&33.11&\ \ 18.67&\ \ 9.48\\
           NOT~\cite{korotin2023neural} &OT& 13.65&\underline{0.9157}&156.07&28.44&\ \ 23.90&15.72\\
           KNOT~\cite{korotin2023kernel} &OT&  \ \ \underline{5.95}&0.8887&118.26&27.33&\ \
           \bf \ \ 3.18&\ \ 9.25\\
           ReFlow~\cite{liu2023flow} &Flow& 70.69&0.4544&\ \ 56.04&29.33&138.59&11.57\\
           EGSDE~\cite{zhao2022egsde} &Diffusion&11.49&0.3835&\ \ 52.11&29.33&\ \ 34.72&11.78\\
           DDIB~\cite{su2023dual} &Diffusion&11.35&0.1275&\ \ 28.45&32.44&\ \ \ \ 9.47&\ \ 9.15\\
           TCR~\cite{mustafa2020transformation}&--&--&--&\ \ 34.61&\underline{40.44}&\ \ \ \ 6.90&\underline{36.21}\\
           SCONES~\cite{daniels2021score} &OT, Diffusion& 15.46&0.1042&\ \ \underline{25.24}&35.33&\ \ \ \ 6.68&10.37\\
           \bf OTCS (ours) &OT, Diffusion& \bf \ \ 1.77&\bf0.9313& \bf\ \ 13.68& \bf96.44& \ \ \ \ \underline{5.12}& \bf67.42\\
        \bottomrule
	\end{tabular}

	\label{tab:quantitative_results}
\end{table}

\subsection{Semi-paired Image-to-Image Translation} \label{sec:semi_paired_i2i}
We consider the semi-paired I2I task that a large number of unpaired  along with a few paired images across source and target domains are given for training. The goal of semi-paired I2I is to leverage the paired cross-domain images to guide the desired translation of source images to the target domain. To apply our approach to the semi-paired I2I, we take the source images as condition data.
We use the $L_2$-regularized semi-supervised OT to estimate the coupling. The cost $c$ and $c'$ in Eq.~\eqref{eq:relation} are taken as the cosine dissimilarity of features using the image encoder of CLIP~\cite{radford2021learning}.
Apart from the compared unpaired I2I approaches in Sect.~\ref{sec:super_resolution}, we additionally compare our approach with the semi-paired I2I approach TCR~\cite{mustafa2020transformation}. 
For the OT-based approaches~\cite{jacob*2019wgan,makkuva2020optimal,rout2022generative,korotin2023neural,korotin2023kernel}, we additionally impose the matching of the paired images using a reconstruction loss in them for semi-paired I2I. It is non-trivial to impose the matching of the paired images in the diffusion/flow-based approaches~\cite{liu2023flow,zhao2022egsde,su2023dual}. 
We adopt two metrics of FID and Accuracy (Acc). The FID measures the quality of generated samples. The higher Acc implies that the guidance of the paired images is better realized for desired translation. The experiments are conducted on digits and natural animal images.

\begin{figure}[t]
\vspace{-0.3cm}
	\centering
        \includegraphics[width=1.0\columnwidth]{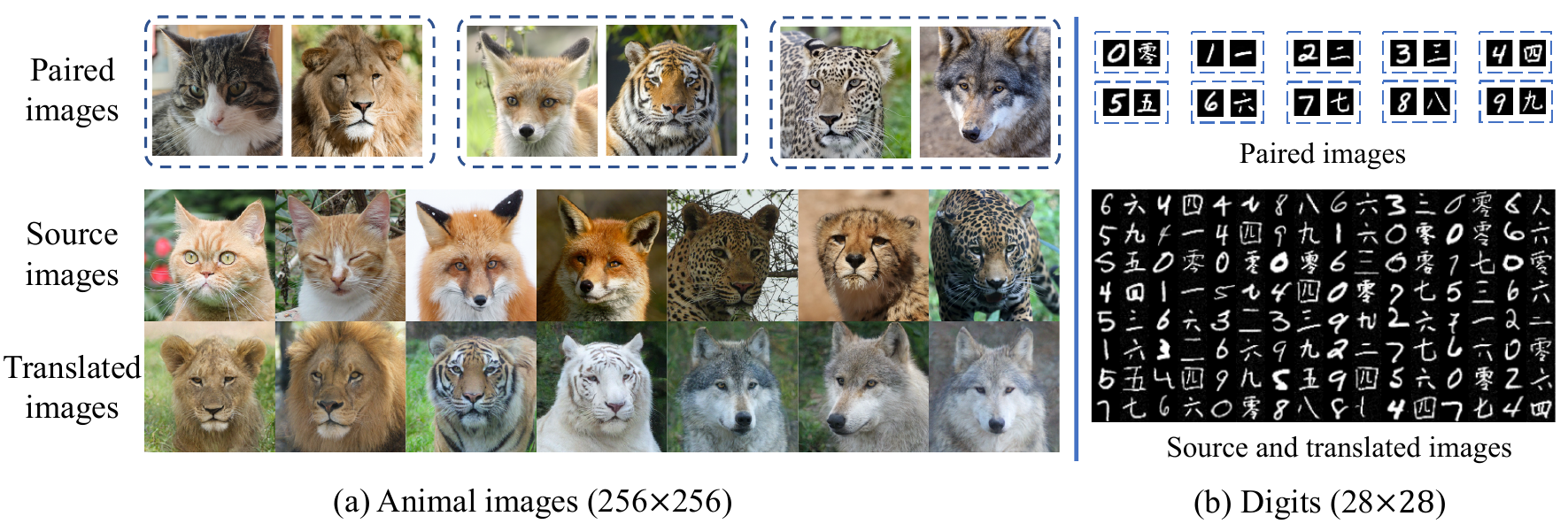}
        \vspace{-0.4cm}
	\caption{Results of OTCS for semi-paired I2I on (a) Animal images and (b) Digits. The bottom of Fig.~\ref{fig:qualitative_afhq}{\color{red}(b)} plots source (odd columns) and corresponding translated (even columns) images.}
    \label{fig:qualitative_afhq}
    \vspace{-0.2cm}
\end{figure}

\begin{figure}[t]
	\centering
        \subfigure[]{\includegraphics[width=0.24\columnwidth]{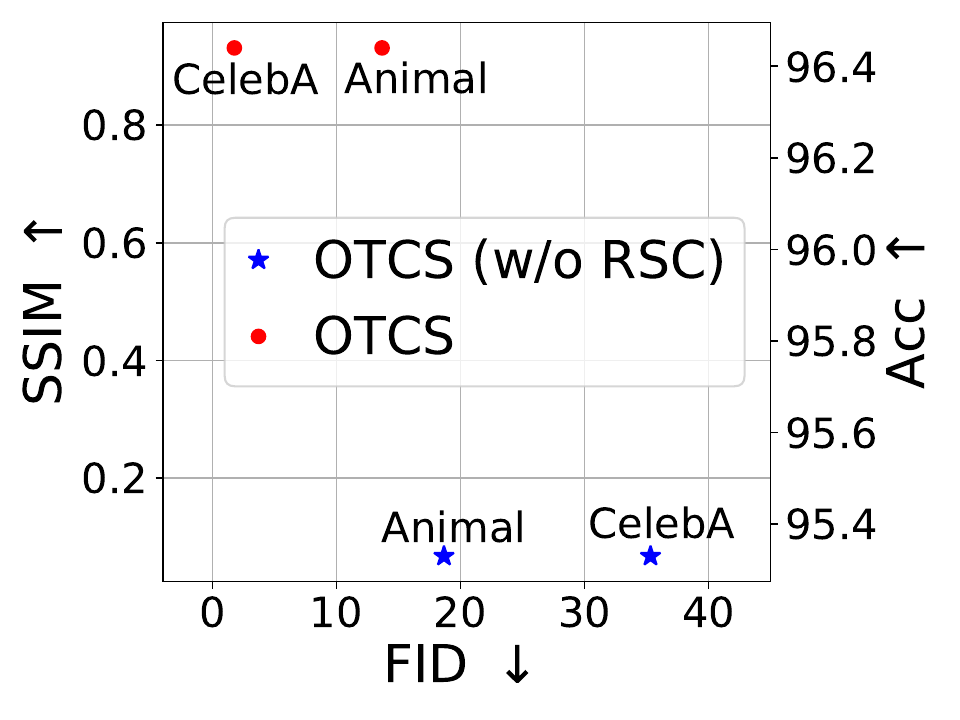}\label{fig:ablation_rsc}}
        \subfigure[]{\includegraphics[width=0.24\columnwidth]{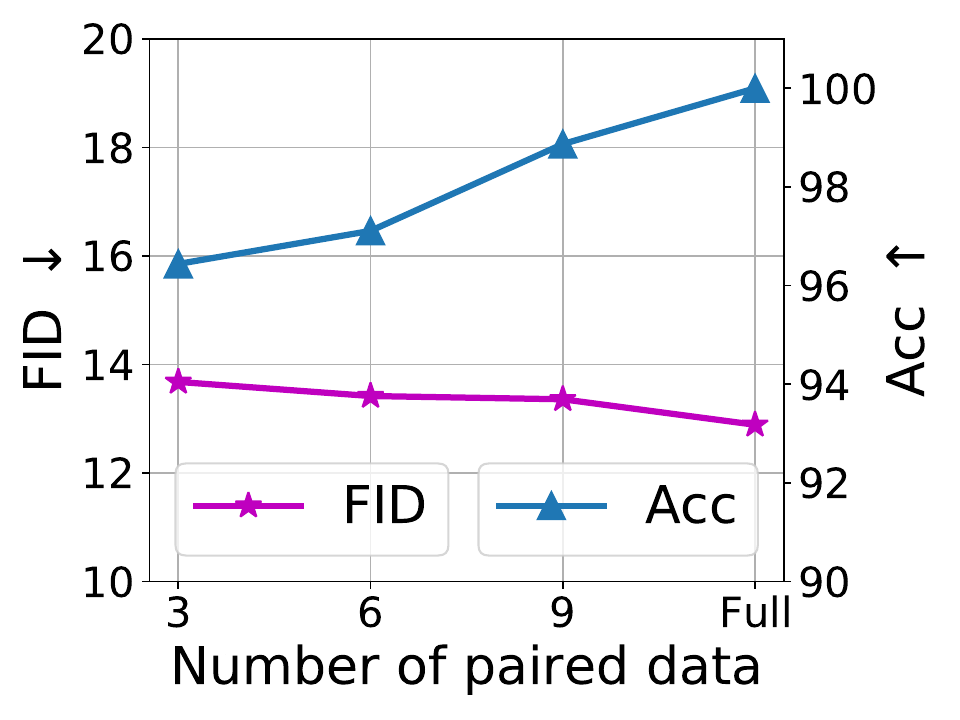}\label{fig:ablation_num}}
        \subfigure[]{\includegraphics[width=0.24\columnwidth]{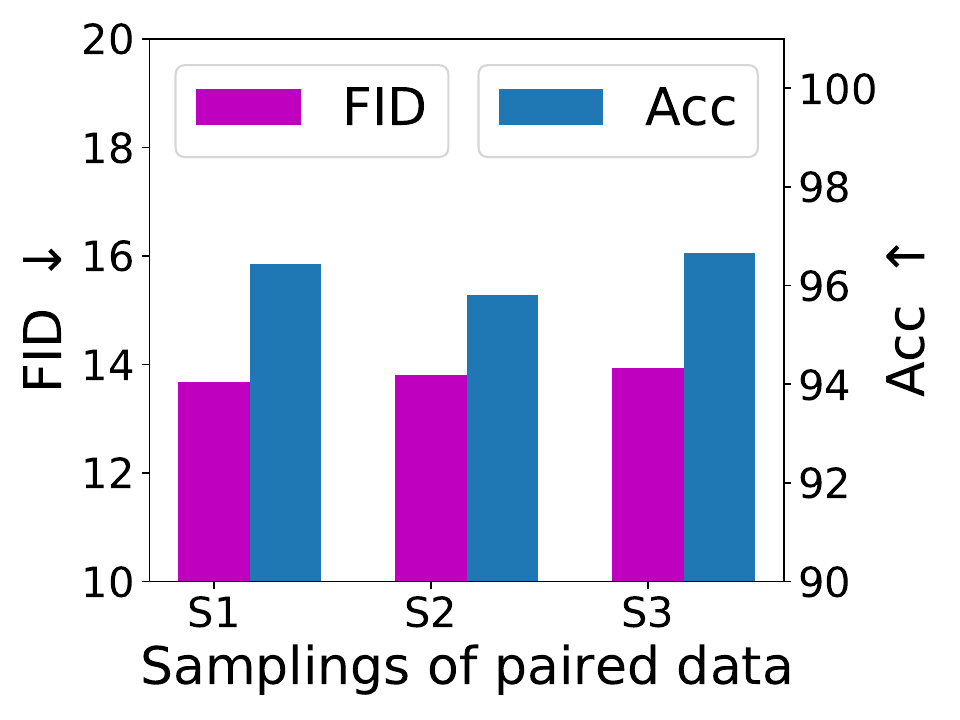}\label{fig:ablation_loc}}
        \subfigure[]{\includegraphics[width=0.24\columnwidth]{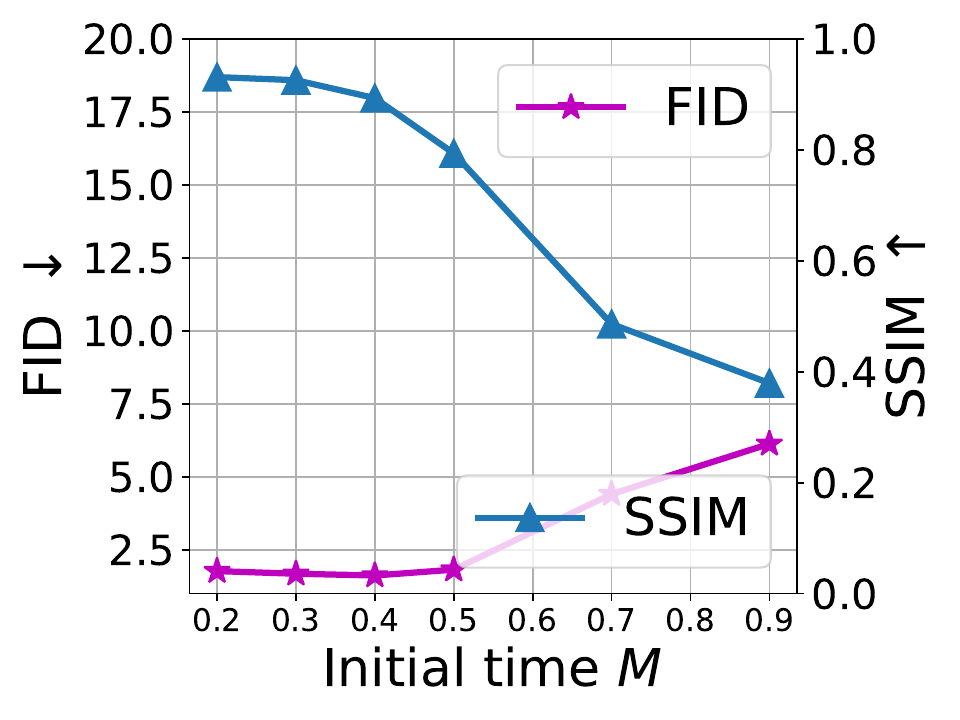}\label{fig:ablation_init}}
        \vspace{-0.3cm}
	\caption{(a) Results of OTCS w/ and w/o RSC. (b-c) Results of OTCS with varying (b) numbers and (c) samplings of paired data on Animal images. (d) Results of OTCS for different $M$ on CelebA.}
    \label{fig:analysis}
    \vspace{-0.2cm}
\end{figure}

For \textit{Animal images}, we take images of cat, fox, and leopard from AFHQ~\cite{choi2020stargan} 
dataset as source, and images of lion, tiger, and wolf  as target. 
Three cross-domain image pairs are given, as shown in Fig.~\ref{fig:qualitative_afhq}{\color{red}(a)}. By the guidance of the paired images, we expect that the cat, fox, and leopard images are respectively translated to the images of lion, tiger, and wolf. The Acc is the ratio of source images translated to ground-truth translated
classes (GTTCs), where the GTTCs of source images of cat/fox/leopard are lion/tiger/wolf. 
For \textit{Digits}, we consider the translation from MNIST~\cite{lecun1998gradient} to Chinese-MNIST~\cite{Chinese-mnist}. 
The MNIST and Chinese-MNIST contain the digits (from 0 to 9) in different modalities.
We annotate 10 cross-domain image pairs, each corresponding to a digit, as in Fig.~\ref{fig:qualitative_afhq}{\color{red}(b)}.  With the guidance of the paired images, we expect that the source images are translated to the target ones representing the same digits. The GTTCs for the source images are their corresponding digits.

We can observe in Tab.~\ref{tab:quantitative_results} that OTCS achieves the highest Acc on both Animal images and Digits, outperforming the second-best method TCR~\cite{mustafa2020transformation} by more than 50\% and 30\% on the two datasets, respectively. 
Our approach explicitly models the guidance of the paired images to the unpaired ones using the semi-supervised OT in training, and better translates the source images to target ones of the desired classes. 
In terms of the FID, OTCS achieves competitive (on Digits) or even superior (on Animal images) results over the other approaches, indicating that OTCS can generate images of comparable quality. The translated images in Figs.~\ref{fig:qualitative_afhq} also show the effectiveness of OTCS.

\subsection{Analysis}
\textbf{Effectiveness of resampling-by-compatibility (RSC).} Figure~\ref{fig:ablation_rsc} shows that OTCS achieves better results (left top points are better) than OTCS (w/o RSC) in semi-paired I2I on Animal images and unpaired super-resolution on CelebA, demonstrating the effectiveness of resampling-by-compatibility. 

\textbf{Results on semi-paired I2I with varying numbers and samplings of paired images.} We study the effect of the number and choice of paired images in semi-paired I2I. Figure~\ref{fig:ablation_num} shows that as the number of paired data increases, the Acc increases, and the FID marginally decreases.
This implies that our approach can impose the guidance of different amounts of paired data to translate source images to desired classes. 
From Fig.~\ref{fig:ablation_loc}, OTCS achieves similar FID and Acc for three different samplings of the same number, \ie, 3, of paired images (denoted as ``S1'', ``S2'', and ``S3'' in Fig.~\ref{fig:ablation_loc}).  

\textbf{Results on unpaired super-resolution with varying initial time.} We show the results of OTCS under different initial time $M$ in the reverse SDE. Figure~\ref{fig:ablation_init} indicates that with a smaller $M$, our OTCS achieves better SSIM. This makes sense because adding smaller-scale noise in inference could better preserve the structure of source data in SBDMs, as in~\cite{meng2021sdedit,zhao2022egsde}. However, smaller $M$ may lead to a larger distribution gap/FID between generated and target data for SBDMs with unconditional score-based model~\cite{meng2021sdedit,zhao2022egsde}. We observe that OTCS achieves the FID below 1.85 for $M$ in [0.2, 0.5]. 


\section{Conclusion}
This paper proposes a novel Optimal Transport-guided Conditional Score-based diffusion model (OTCS) for image translation with unpaired or partially paired training dataset. 
We build the coupling of the condition data and target data using $L_2$-regularized unsupervised and semi-supervised OTs, and present the OT-guided conditional denoising score-matching and resampling-by-compatibility to train the conditional score-based model. 
Extensive experiments in unpaired super-resolution and semi-paired I2I tasks demonstrated the effectiveness of OTCS for achieving desired translation of source images. 
We theoretically analyze that OTCS realizes data transport for OT.
In the future, we are interested in more applications of OTCS, such as medical image translation/synthesis.

\section*{Limitations}
The cost function should be determined first when using our method. In experiments, we simply choose the squared $L_2$-distance in image space for unpaired super-resolution and cosine distance in feature space for semi-paired I2I, achieving satisfactory performance. However, the performance may be improved if more domain knowledge is employed to define the cost function.  Meanwhile, if the number of target data is small, the generation ability of our trained model may be limited. 

\section*{Acknowledgement}
This work was supported by National Key R\&D Program 2021YFA1003002 and NSFC (12125104, U20B2075, 61721002).

\bibliographystyle{unsrt}
\bibliography{egbib}




\newcommand{\mf}{\fontsize{8.5pt}{\baselineskip}\selectfont}
\renewcommand\thesection{Appendix \Alph{section}}
\renewcommand{\thetable}{A-\arabic{table}}
\renewcommand{\theequation}{A-\arabic{equation}}
\renewcommand{\thedefinition}{A-\arabic{definition}}
\renewcommand{\thefigure}{A-\arabic{table}}
\renewcommand{\thefigure}{A-\arabic{figure}}

\setcounter{proposition}{0}
\setcounter{equation}{0}
\setcounter{figure}{0}
\setcounter{table}{0}
\setcounter{thm}{0}
\newpage
{\centering \Large{\bf Appendix}}
\appendix
\section{Additional Details for Sections 2 and 3}\label{sec:app_A}
\subsection{Additional Details for Section 2}
\paragraph{Details of VP-SDE and VE-SDE.}
As mentioned in Sect.~2.1, we choose the VE-SDE and the VP-SDE as the forward SDEs. For VE-SDE, $f(\mathbf{y},t)=0$ and $g(t)=\alpha^t$ where $\alpha$ is a hyper-parameter. For VP-SDE, $f(\mathbf{y},t)=-\frac{1}{2}\beta(t)\mathbf{y}$ and $g(t)=\sqrt{\beta(t)}$ where $\beta(t)=\beta_{\rm min}+(\beta_{\rm max}-\beta_{\rm min})t$, and $\beta_{\rm min}$ and $\beta_{\rm max}$ are hyper-parameters.
Then, the conditional distribution, \textit{a.k.a.}, permutation kernel, $p_{t|0}(\mathbf{y}_t|\mathbf{y}_0)$ of $\mathbf{y}_t$ given $\mathbf{y}_0$ is
\begin{equation}\label{eq_a:conditional_distribution_noise}
    p_{t|0}(\mathbf{y}_t|\mathbf{y}_0) = 
    \begin{cases}
    \mathcal{N}\left(\mathbf{y}_t|{\bf \mathbf{y}}_0,\frac{1}{2\log\alpha}(\alpha^{2t}-1)\bf{I})\right), & \mbox{ for VE-SDE, }\\
    \mathcal{N}\left(\mathbf{y}_t|\mathbf{y}_0e^{\frac{1}{2}h(t)},(1-e^{h(t)}){\bf I} \right), & \mbox{ for VP-SDE, }
    \end{cases}
\end{equation}
 where $h(t)=-\frac{1}{2}t^2(\beta_{\rm max}-\beta_{\rm min})-t\beta_{\rm min}$, and $\bf I$ is the identity matrix. Following~\cite{song2020score}, we set $T$ to $1$, and $p_{\rm prior}=\mathcal{N}(0,\mathbf{I})$ for VP-SDE and $p_{\rm prior}=\mathcal{N}(0,\frac{1}{2\log\alpha}(\alpha^{2t}-1)\mathbf{I})$ for VE-SDE.

\paragraph{Pseudo-codes of algorithm for training ${u_{\omega},v_{\omega}}$.} The pseudo-codes of the algorithm to learn the dual variables $u_{\omega},v_{\omega}$, \textit{a.k.a.}, potentials, are given in Algorithm~\ref{alg:alg0}.

\begin{algorithm}[H]\label{alg:alg0}
  \caption{Algorithm for estimating potentials $u_{\hat{\omega}},v_{\hat{\omega}}$}
  \SetAlgoLined
  \KwIn{Distribution $p$ of conditions, target data distribution $q$, paired data (if available)
  }
  \KwOut{Learned potentials $u_{\hat{\omega}},v_{\hat{\omega}}$}
  \For{$ {\rm iter} = 1,\cdots,N_{\rm iter}'$}{
    Sampling mini-batch data $\bm{X}=\{\x_b\}_{b=1}^{B'}$ from $p$, $\bm{Y}=\{\y_b\}_{b=1}^{B'}$ from $q$\;
    \eIf{paired data are available}
    {Computing the loss of semi-supervised OT in Eq.~(6) on $\bm{X}$ and $\bm{Y}$\;}
    {Computing the loss of unsupervised OT in Eq.~(6) on $\bm{X}$ and $\bm{Y}$\;}
    Backward propagation to compute the gradient and update $\omega$ using Adam algorithm\;
    }
  $\hat{\omega} = \omega$. 
\end{algorithm}

\subsection{Additional Details for Section 3}
\paragraph{Rationality of the resampling-by-compatibility.}
We next explain the rationality of the resampling-by-compatibility presented in Sect.~3.3. For the convenience of description, for any $\x,\y$, we denote 
\begin{equation}\label{eq_a:loss_for_xy}
    \mathcal{J}_{\x,\y}=\mathbb{E}_t w_t\mathbb{E}_{\y_t\sim p_{t|0}(\y_t|\y)}\Vert s_{\theta}(\y_t;\x,t) -\nabla_{\y_t} \log p_{t|0}(\y_t|\y)\Vert_2^2.
\end{equation}
The training loss $\mathcal{J}_{\rm CDSM}(\theta)$ in Eq.~(9) can be written as 
\begin{equation}\label{eq_a:j_xy}
   \mathcal{J}_{\rm CDSM}(\theta) = {\E}_{\x\sim p}{\mathbb{E}}_{\y\sim q}H(\x,\y)\mathcal{J}_{\x,\y}. 
\end{equation}
 By the resampling-by-compatibility, $q$ is approximated based on samples $\mathbf{Y}_{\x}$ by $q(\y)\approx \frac{1}{L}\sum_{l=1}^L \delta(\y-\y^l)$.
We then have
\begin{equation}\label{eq_a:equation_approx}
    \begin{split}
        \mathcal{J}_{\rm CDSM}(\theta) 
        &\approx {\E}_{\x\sim p} \frac{1}{L}\sum_{l=1}^L H(\x,\y^l)\mathcal{J}_{\x,\y^l}\\
        &\propto {\E}_{\x\sim p} \frac{1}{H_0}\sum_{l=1}^L H(\x,\y^l)\mathcal{J}_{\x,\y^l} \\
        &={\E}_{\x\sim p}\E_{\y\sim \tilde{h}_{\x}}\mathcal{J}_{\x,\y},
    \end{split}
\end{equation}
where $H_0 = \sum_{l=1}^L H(\x,\y^l)$ and
$\tilde{h}_{\x}$ is the distribution defined on $\bm{Y}_{\x}$ as 
$\tilde{h}_{\x}(\y)=\frac{1}{H_0}\sum_{l=1}^L H(\x,\y^l)\delta(\y-\y^l)$. Equation~\eqref{eq_a:equation_approx} indicates that $\mathcal{J}_{\rm CDSM}(\theta)$ can be approximately implemented based on samples using our resampling-by-compatibility strategy. More concretely, the last line in Eq.~\eqref{eq_a:equation_approx} can be implemented by sequentially dropping $\x$ from $p$, generating samples $\bm{Y}_{\x}$ to construct $\tilde{h}_{\x}$, sampling $\y$ from $\tilde{h}_{\x}$, and computing $\mathcal{J}_{\x,\y}$ on $(\x,\y)$. Note that dropping sample $\y$ from $\tilde{h}_{\x}$ means to choose a $\y$ from $\bm{Y}_{\x}$ with the probability proportional to $H(\x,\y^l)$. 

 
\paragraph{Training by fitting noise.} Based on Eq.~\eqref{eq_a:conditional_distribution_noise}, $p_{t|0}(\y_t|\y)$ is a Gaussian distribution. We denote the $\sigma_t\mathbf{I}$ as the standard variation of $p_{t|0}(\y_t|\y)$, \ie, $\sigma_t^2=\frac{1}{2\log \alpha}(\alpha^{2t}-1)$ for VE-SDE, and $\sigma_t^2=1-e^{h(t)}$ for VP-SDE. Using the reparameterization trick~\cite{kingma2013auto}, given $\x, \y$ sampled using our resampling-by-compatibility, we have $\y_t = \y + \sigma_t \bm{\epsilon}$ for VE-SDE, and $\y_t = e^{\frac{1}{2}h(t)}\y + \sigma_t \bm{\epsilon}$ for VP-SDE, where $\bm{\epsilon}\sim \mathcal{N}(0,\mathbf{I})$.
Further, $\nabla_{\y_t}\log p_{t|0}(\y_t|\y)=-\frac{1}{\sigma_t}\bm{\epsilon}$. Therefore, the loss $\mathcal{J}_{\x,\y}$ in Eq.~\eqref{eq_a:loss_for_xy} can be written as 
\begin{equation}\label{eq_a:conditional_denoising_score_matching2} 
    \mathcal{J}_{\x,\y}  = 
    {\mathbb{E}}_{t, \bm{\epsilon}\sim \mathcal{N}(0,\mathbf{I})}\left[\frac{w_t}{\sigma_t^2}\left\Vert s_{\theta}(\bm{u}_t(\y)+\sigma_t\bm{\epsilon};\x,t)\sigma_t + \bm{\epsilon}\right\Vert_2^2\right].
\end{equation} 
where $\bm{u}_t(\y) = \y$ for VE-SDE, and $\bm{u}_t(\y) = e^{\frac{1}{2}h(t)}\y$ for VP-SDE. Equation~\eqref{eq_a:conditional_denoising_score_matching2} implies that $s_{\theta}(\y_t;\x,t)$ is trained to fit the scaled noise $-\frac{1}{\sigma_t}\bm{\epsilon}$.

\paragraph{Pseudo-codes of algorithm training ${s_{\theta}}$.}The pseudo-codes of the algorithm to train $s_{\theta}$ for the case where training data consist of condition dataset $\mathcal{D}_{\x}$ and target dataset $\mathcal{D}_{\y}$ are given in Algorithm~\ref{alg:alg2}.  The pseudo-codes of the algorithm to train $s_{\theta}$ for the case with continuous distributions $p,q$ are given in Algorithm~\ref{alg:alg1}. 
\begin{algorithm}[t]\label{alg:alg2}
  \caption{Training algorithm for discrete datasets}
  \KwIn{Condition dataset $\mathcal{D}_{\x}$, target dataset $\mathcal{D}_{\y}$, paired data (if available)}
  \KwOut{Trained conditional score-based model $s_{\hat{\theta}}$}
  Learning potentials $u_{\hat{\omega}},v_{\hat{\omega}}$ using Algorithm~\ref{alg:alg0}\;
  \tcp{Computing and storing $H$ for target samples with non-zero $H$}
  $Dict = \{\}$\;
  \For{$\x$ in $\mathcal{D}_{\x}$}{
  $\bm{Y}_{\x}=\{\y:H(\x,\y)>0, \y\in\mathcal{D}_{\y}\}$\;
  $\bm{H}_{\x} = \{\frac{H(\x,\y)}{H_0}:\y \in \bm{Y}_{\x}\}$, where $H_0 = \sum_{\y\in\bm{Y}_{\x}}H(\x,\y)$\;
  $Dict = Dict \cup \{(\bm{Y}_{\x},\bm{H}_{\x})\}$\;
  }
  \tcp{Traning $s_{\theta}$ on mini-batch data}
  \For{$ {\rm iter} = 1,\cdots,N_{\rm iter}$}{
    Sampling mini-batch data $\{\x_b\}_{b=1}^B$ from $\mathcal{D}_{\x}$\;
    \For{$b=1,2,\cdots,B$}
       {\tcp{Resampling-by-compatibility}
        Finding $(\bm{Y}_{\x_b},\bm{H}_{{\x}_b})$ in $Dict$ \;
        Choosing $\y_b$ from $\bm{Y}_{\x_b}$ with probability $\bm{H}_{{\x}_b}$\;
        Sampling $t_b$ from $\mathcal{U}([0,T])$, and $\bm{\epsilon}_b$ from $\mathcal{N}(0,\mathbf{I})$\;
       }
       Computing loss $\frac{1}{B}\sum_{b=1}^B\frac{w_{t_b}}{\sigma_{t_b}^2}\left\Vert s_{\theta}\left(\bm{u}_{t_b}(\y_b)+\sigma_{t_b}\bm{\epsilon}_b;\x_b,t_b\right) \sigma_{t_b} + \bm{\epsilon}_{b} \right\Vert_2^2$ \tcp*{ Eq.~\eqref{eq_a:conditional_denoising_score_matching2}}
       Backward propagation to compute the gradient \textit{w.r.t.} $\theta$ and update $\theta$ using Adam algorithm\;
    }
    $\hat{\theta} = \theta $. 
\end{algorithm}

\begin{algorithm}[t]\label{alg:alg1}
  \caption{Training algorithm for continuous distributions}
  \SetAlgoLined
  \KwIn{Condition distribution $p$, data distribution $q$, paired data (if available)}
  \KwOut{Trained conditional score-based model $s_{\hat{\theta}}$}
  Learning potentials $u_{\hat{\omega}},v_{\hat{\omega}}$ using Algorithm~\ref{alg:alg0}\;
  \tcp{Traning $s_{\theta}$ on mini-batch data}
  \For{$ {\rm iter} = 1,\cdots,N_{\rm iter}$}{
    Sampling mini-batch data $\{\x_b\}_{b=1}^B$ from $p$\;
    \For{$b=1,2,\cdots,B$}
       {\tcp{Resampling-by-compatibility}
       Sampling $\bm{Y}_{\x}=\{\y^l\}_{l=1}^L$ from $q$\;
        Computing $h^l=H(\x,\y^l)$ for all $l$ as in Eq.~(7)\;
        Choosing $\y_b$ from $\bm{Y}_{\x}$ with probability $\frac{1}{\sum_{l=1}^Lh^l}(h^1,h^2,\cdots,h^L)$\;
        Sampling $t_b$ from $\mathcal{U}([0,T])$, and $\bm{\epsilon}_b$ from $\mathcal{N}(0,\mathbf{I})$\;
       }
    Computing loss $\frac{1}{B}\sum_{b=1}^B\frac{w_{t_b}}{\sigma_{t_b}^2}\left\Vert s_{\theta}\left(\bm{u}_{t_b}(\y_b)+\sigma_{t_b}\bm{\epsilon}_b;\x_b,t_b\right) \sigma_{t_b} + \bm{\epsilon}_{b} \right\Vert_2^2$ \tcp*{ Eq.~\eqref{eq_a:conditional_denoising_score_matching2}}
       Backward propagation to compute the gradient \textit{w.r.t.} $\theta$ and update $\theta$ using Adam algorithm\;
    }
    $\hat{\theta} = \theta $. 
\end{algorithm}

\section{Proofs}
\subsection{Proof of Proposition 1}
\begin{proposition}\label{lem_a:lemma1}
Let $\mathcal{C}(\x,\y)=\frac{1}{p(\x)}\delta(\x-\x_{\rm cond}(\y))$ where $\delta$ is the Dirac delta function, then $\mathcal{J}_{\rm DSM}(\theta)$ in Eq.~(1) can be reformulated as
\begin{equation}\label{eq_a:reformulation_paired_data}
    \mathcal{J}_{\rm DSM}(\theta)=\E_tw_t\E_{\x\sim p}\E_{\y\sim q}\mathcal{C}(\x,\y) \E_{\y_t\sim p_{t|0}(\y_t|\y)}
    \left\Vert s_{\theta}(\y_t;\x,t) - \nabla_{\y_t}\log p_{t|0}(\y_t|\y)\right\Vert_2^2.
\end{equation}
Furthermore, $\gamma(\x,\y)=\mathcal{C}(\x,\y)p(\x)q(\y)$ is a joint distribution for marginal distributions $p$ and $q$. 
\end{proposition}

\paragraph{Proof.} We first i) prove Eq.~\eqref{eq_a:reformulation_paired_data}, and then ii) show that $\gamma(\x,\y)$ is a joint distribution for marginal distributions $p$ and $q$.

i) The right side of Eq.~\eqref{eq_a:reformulation_paired_data} is 
\begin{equation*}
\begin{split}
       &\E_tw_t\E_{\x\sim p}\E_{\y\sim q}\mathcal{C}(\x,\y) \E_{\y_t\sim p_{t|0}(\y_t|\y)}\left\Vert s_{\theta}(\y_t;\x,t) - \nabla_{\y_t}\log p_{t|0}(\y_t|\y)\right\Vert_2^2\\
       =&\E_tw_t\E_{\y\sim q}\int p(\x)\mathcal{C}(\x,\y) \E_{\y_t\sim p_{t|0}(\y_t|\y)}\left\Vert s_{\theta}(\y_t;\x,t) - \nabla_{\y_t}\log p_{t|0}(\y_t|\y)\right\Vert_2^2\dif \x\\
\end{split}
\end{equation*}
\begin{equation*}
\begin{split}
       =&\E_tw_t\E_{\y\sim q}\int \delta(\x-\x_{\rm cond}(\y)) \E_{\y_t\sim p_{t|0}(\y_t|\y)}\left\Vert s_{\theta}(\y_t;\x,t) - \nabla_{\y_t}\log p_{t|0}(\y_t|\y)\right\Vert_2^2\dif \x\\
       =&\E_tw_t\E_{\y\sim q}\E_{\y_t\sim p_{t|0}(\y_t|\y)}\left\Vert s_{\theta}(\y_t;\x_{\rm cond}(\y),t) - \nabla_{\y_t}\log p_{t|0}(\y_t|\y)\right\Vert_2^2,
\end{split}
\end{equation*}
which is the definition of $\mathcal{J}_{\rm DSM}(\theta)$ in Eq.~(1).

ii) We show that the marginal distributions of $\gamma(\x,\y)$ are respectively $p$ and $q$ as follows. Firstly,
\begin{equation}
\int \gamma(\x,\y) \dif \x = \int \delta(\x-\x_{\rm cond}(\y)) q(\y)\dif\x =q(\y)\int \delta(\x-\x_{\rm cond}(\y)) \dif\x = q(\y).
\end{equation}
Secondly, from the definition of $\delta(\cdot)$, we have $\delta(\x-\x_{\rm cond}(\y))=\sum_{\{\y':\x_{\rm cond}(\y')=\x\}}\delta(\y-\y')$. Then, we have
\begin{equation}
\begin{split}
        \int \gamma(\x,\y) \dif \y = &\int \delta(\x-\x_{\rm cond}(\y)) q(\y)\dif\y\\
        = &\int \sum_{\{\y':\x_{\rm cond}(\y')=\x\}}\delta(\y'-\y) q(\y)\dif\y\\
        =&\sum_{\{\y':\x_{\rm cond}(\y')=\x\}}\int\delta(\y'-\y) q(\y)\dif\y\\
        =&\sum_{\{\y':\x_{\rm cond}(\y')=\x\}}q(\y') \\
        =& p(\x).
\end{split}
\end{equation}

\subsection{Proof of Theorem 1}
\begin{thm}\label{thm_a:thm1}
        For $\x\sim p$, we define the forward SDE $\dif \y_t = f(\y_t,t)\dif t + g(t)\dif\mathbf{w}$ with $\y_0\sim\hat{\pi}(\cdot|\x)$ and $t\in [0,T]$, where $f,g,T$ are given in Appendix A.1. Let $p_{t}(\y_t|\x)$ be the corresponding distribution of $\y_t$ and
        $\mathcal{J}_{\rm CSM}(\theta) =  \mathbb{E}_{t}w_t\mathbb{E}_{\x\sim p}\mathbb{E}_{\y_t\sim p_t(\y_t|\x)} \left\Vert s_{\theta}(\mathbf{y}_t;\x,t) - \nabla_{\y_t}\log p_t(\y_t|\x)\right\Vert_2^2$,
    then
     we  have  $\nabla_{\theta}\mathcal{J}_{\rm CDSM}(\theta)=\nabla_{\theta}\mathcal{J}_{\rm CSM}(\theta)$.
\end{thm}

\paragraph{Proof.} Given $\x$ and $t$, $p_t(\y_t|\x)$ is the distribution of $\y_t$ produced by the forward SDE $\dif \y_t = f(\y_t,t)\dif t + g(t)\dif\mathbf{w}$ with initial state $\y_0\sim\hat{\pi}(\y_0|\x)$. This implies that $\x\rightarrow\y_0\rightarrow\y_t$ is a Markov Chain. So the distribution $p_{t|0}(\y_t|\y_0,\x)$ of $\y_t$ given $\y_0$ and $\x$ depends on $\y_0$ but $\x$, \ie, $p_{t|0}(\y_t|\y_0,\x)=p_{t|0}(\y_t|\y_0)$, where $p_{t|0}(\y_t|\y_0)$ is the distribution of $\y_t$ by the forward SDE $\dif \y_t = f(\y_t,t)\dif t + g(t)\dif\mathbf{w}$ with initial state $\y_0$. 
According to~\cite{vincent2011connection}, given any $\x$ and $t$, we have 
\begin{equation}
    \begin{split}
    &\E_{\y_0\sim\hat{\pi}(\y_0|\x)}\mathbb{E}_{\y_t\sim p_{t|0}(\y_t|\y_0,\x)}\Vert s_{\theta}(\y_t;\x,t) -\nabla_{\y_t} \log p_{t|0}(\y_t|\y_0,\x)\Vert_2^2 \\
    = &\mathbb{E}_{\y_t\sim p_t(\y_t|\x)} \left\Vert s_{\theta}(\mathbf{y}_t;\x,t) - \nabla_{\y_t}\log p_t(\y_t|\x)\right\Vert_2^2 + C_{\x,t},\\
    \end{split}
\end{equation}
where $C_{\x,t}$ is a constant to $\theta$ depending on $\x$ and $t$. Then, we have 
\begin{equation}
    \begin{split}
    &w_t\left(\mathbb{E}_{\y_t\sim p_t(\y_t|\x)} \left\Vert s_{\theta}(\mathbf{y}_t;\x,t) - \nabla_{\y_t}\log p_t(\y_t|\x)\right\Vert_2^2 + C_{\x,t}\right)\\
    =&w_t\E_{\y_0\sim\hat{\pi}(\y_0|\x)}\mathbb{E}_{\y_t\sim p_{t|0}(\y_t|\y_0)}\Vert s_{\theta}(\y_t;\x,t) -\nabla_{\y_t} \log p_{t|0}(\y_t|\y_0)\Vert_2^2. \\
    \end{split}
\end{equation}
Taken expectation over $\x$ and $t$ in the above equation, we have 
\begin{equation}
\begin{split}
        &\mathcal{J}_{\rm CSM}(\theta) + \E_{\x\sim p}\E_tw_tC_{\x,t} \\
    = & \E_tw_t\E_{\x\sim p}\E_{\y_0\sim\hat{\pi}(\y_0|\x)}\mathbb{E}_{\y_t\sim p_{t|0}(\y_t|\y_0)}\Vert s_{\theta}(\y_t;\x,t) -\nabla_{\y_t} \log p_{t|0}(\y_t|\y_0)\Vert_2^2\\
    = & \E_tw_t\E_{\x\sim p}\E_{\y_0\sim q}H(\x,\y_0)\mathbb{E}_{\y_t\sim p_{t|0}(\y_t|\y_0)}\Vert s_{\theta}(\y_t;\x,t) -\nabla_{\y_t} \log p_{t|0}(\y_t|\y_0)\Vert_2^2\\
    = &\mathcal{J}_{\rm CDSM}(\theta).
\end{split}
\end{equation}
Since $\E_{\x\sim p}\E_tw_tC_{\x,t}$ is a constant to $\theta$, 
we have 
\begin{equation}
    \nabla_{\theta}\mathcal{J}_{\rm CDSM}(\theta)=\nabla_{\theta}\mathcal{J}_{\rm CSM}(\theta).
\end{equation}
\subsection{Assumptions and Proof of Theorem~2}
\begin{thm}\label{thm_a:bound}
Suppose the assumptions in Appendix B.3.1 hold, and $w_t=g(t)^2$, then we have 
\begin{equation}\label{eq_a:bound}
\begin{split}
    \E_{\x\sim p}W_2\left(p^{\rm sde}(\cdot|\x),\pi(\cdot|\x)\right)
 \leq & C_1 \big\Vert \nabla_{\hat{\pi}}\mathcal{L} (\hat{\pi}, u_{\hat{\omega}}, v_{\hat{\omega}})\big\Vert_1+ \sqrt{C_2 \mathcal{J}_{\rm CSM}(\hat{\theta})} \\
  +& C_3\E_{\x\sim p}W_2\left(p_T(\cdot|\x),p_{\rm prior}\right),
\end{split}
\end{equation}
where $C_1,C_2$, and $C_3$ are constants to $\hat{\omega}$ and $\hat{\theta}$.
\end{thm}
\subsubsection{Assumptions}
\begin{enumerate}[(1)]
    \item $f(\y,t)$ is Lipschitz continuous in the space variable $\y$: there exists a positive constant $L_f(t)\in(0,\infty)$ depending on $t\in[0,T]$ such that for all $\y_1,\y_2$,
    \begin{equation}
         \Vert f(\y_1,t)-f(\y_2,t)\Vert_2\leq L_f(t) \Vert \y_1-\y_2\Vert_2.
    \end{equation}
    
    \item $s_{\theta}(\y;\x,t)$ satisfies the one-sided Lipschitz condition: there exists a constant $L_s(t)$ depending on $t$, such that for all $\y_1,\y_2$,
        \begin{equation}
            (s_{\theta}(\y_1;\x,t)-s_{\theta}(\y_2;\x,t))(\y_1-\y_2)\leq L_s(t) \Vert \y_1-\y_2\Vert_2,
        \end{equation}
    for any $\x$.
    \item For any $\x$, $
    \E_{\hat{\pi}(\cdot|\x)}[|\log\hat{\pi}(\cdot|\x)|], \E_{\hat{\pi}(\cdot|\x)}[\log |\Lambda(\y)|],\E_{p_{\rm prior}}[|\log p_{\rm prior}|], \mbox{ and }   \E_{p_{\rm prior}}[\log |\Lambda(\y)|]$ are finite,
    where $\Lambda(\y) = \log\max(\Vert \y\Vert_2,1)$.

    \item There exists positive constants $A_1$ and $A_2$ such that
    \begin{equation}
        f(\y,t)\y\leq A_1\Vert \y\Vert_2 +A_2, \forall \y, \forall t\in[0,T].
    \end{equation}

    \item There exists a positive constant $A_3$ such that
    \begin{equation}
        \frac{1}{A_3}<g(t)<A_3, \forall t\in[0,T].
    \end{equation}

    \item $\int_0^T\E_{p_t(\cdot|\x)}[f^2]\dif t$, $\int_0^T\E_{q_t(\cdot|\x)}[(f-g^2s_{\theta})] \dif t$ are finite for any $\x$, where $q_t(\y_t|\x)$ is the distribution produced by the reverse SDE in Eq.~(10) at time $t$.

    \item  $\hat{\pi}(\cdot|\x), p_{\rm prior}$ are in $C^2$ \textit{w.r.t.} $\y$ for any $\x$. $f,g,s_{\theta}$ are in $C^2$ \textit{w.r.t.} $\y$ and $t$ for any $\x$.

    \item There exists $k > 0$ such that $p_t(\y|\x) = \mathcal{O}(\exp(-\Vert \y\Vert_2^k))$ and $q_t(\y|\x) = \mathcal{O}(\exp(-\Vert \y\Vert_2^k))$ for any $ t \in
[0,T]$ and any $\x$. 

    \item $\mathcal{L} (\pi,u,v)$ is $\kappa$-strongly convex in $L_1$-norm \textit{w.r.t.} $\pi$.
\end{enumerate}
Assumptions (1)-(8) are based on the assumptions in~\cite{kwon2022scorebased} that investigates the bound for unconditional SBDMs. For Assumption (9), $\mathcal{L}(\pi,u,v)$ is strongly convex as proved in~\cite{daniels2021score}.

\subsubsection{Proof}
Since $W_2(\cdot,\cdot)$ is a proper metric, using the triangle inequality, we have
\begin{equation}\label{eq_a:triangle}
    \E_{\x\sim p}W_2\left(p^{\rm sde}(\cdot|\x),\pi(\cdot|\x)\right)\leq \E_{\x\sim p}W_2\left(p^{\rm sde}(\cdot|\x),\hat{\pi}(\cdot|\x)\right)+\E_{\x\sim p}W_2\left(\hat{\pi}(\cdot|\x),\pi(\cdot|\x)\right).
\end{equation}
We next respectively bound the right-side terms.
\paragraph{Bounding $\bm{\E_{\x\sim p}W_2\left(p^{\rm sde}(\cdot|\x),\hat{\pi}(\cdot|\x)\right)}$.} 
Let $I(t)=\exp\left(\int_0^tL_f(r)+L_s(r)g(r)^2\dif r\right)$.
According to Corollary 1 in~\cite{kwon2022scorebased}, for any $\x$, we have 
\begin{equation}\label{eq_a:sm_bound}
    W_2\left(p^{\rm sde}(\cdot|\x),\hat{\pi}(\cdot|\x)\right) \leq
\sqrt{T\left(\int_0^Tg(t)^2I(t)^2\dif t\right)\mathcal{J}_{\rm SM}^{\x}(\hat{\theta})} + I(T)W_2\left(p_T(\cdot|\x),p_{\rm prior}\right),
\end{equation}
where $\mathcal{J}^{\x}_{\rm SM}(\hat{\theta})= \E_tw_t\E_{\y_t\sim p_t(\y|\x)}\Vert s_{\hat{\theta}}(\y_t;\x,t)-\nabla_{\y_t}\log p_t(\y_t|\x)\Vert_2^2$.
Taking expectation over $\x$ in Eq.~\eqref{eq_a:sm_bound}, we have 
\begin{equation}
\begin{split}
    \E_{\x\sim p}W_2\left(p^{\rm sde}(\cdot|\x),\hat{\pi}(\cdot|\x)\right) &\leq
\E_{\x\sim p}\left(\sqrt{T\left(\int_0^Tg(t)^2I(t)^2\dif t\right)\mathcal{J}_{\rm SM}^{\x}(\hat{\theta})}\right)\\& + I(T)\E_{\x\sim p}W_2\left(p_T(\cdot|\x),p_{\rm prior}\right).
\end{split}
\end{equation}
Since $\sqrt{x}$ is concave in $[0,\infty)$, using the Jesen-Inequality, we have $\E[\sqrt{x}]\leq \sqrt{\E[x]}$. Therefore, 
\begin{equation}
\begin{split}
    &\E_{\x\sim p}W_2\left(p^{\rm sde}(\cdot|\x),\hat{\pi}(\cdot|\x)\right) \\&\leq
\sqrt{T\left(\int_0^Tg(t)^2I(t)^2\dif t\right)\E_{\x\sim p}[\mathcal{J}_{\rm SM}^{\x}(\hat{\theta})]} + I(T)\E_{\x\sim p}W_2\left(p_T(\cdot|\x),p_{\rm prior}\right)\\
&=\sqrt{T\left(\int_0^Tg(t)^2I(t)^2\dif t\right)\mathcal{J}_{\rm CSM}(\hat{\theta})} + I(T)\E_{\x\sim p}W_2\left(p_T(\cdot|\x),p_{\rm prior}\right).
\end{split}
\end{equation}
Let $C_2 = T\left(\int_0^Tg(t)^2I(t)^2\dif t\right)$ and $C_3 = I(T)$. Then, we have 
\begin{equation}\label{eq_a:bound_c2_c3}
    \E_{\x\sim p}W_2\left(p^{\rm sde}(\cdot|\x),\hat{\pi}(\cdot|\x)\right)\leq \sqrt{C_2\mathcal{J}_{\rm CSM}(\hat{\theta})} + C_3\E_{\x\sim p}W_2\left(p_T(\cdot|\x),p_{\rm prior}\right).
\end{equation}
\paragraph{Bounding $\bm{\E_{\x\sim p}W_2\left(\hat{\pi}(\cdot|\x),\pi(\cdot|\x)\right)}$.} 
According to Remark 2.26 in~\cite{peyre2019computational} (the relation between the Wasserstein distance and $L_1$-distance), we have 
\begin{equation}
    W_2(\mu,\nu) \leq \max_{\x,\y \in \mathcal{X}}\{\Vert \x- \y\Vert_2\}\Vert\mu-\nu\Vert_1,
\end{equation}
for any $\mu,\nu$ supported on $\mathcal{X}$.
We then have 
\begin{equation}
    W_2\left(\hat{\pi}(\cdot|\x),\pi(\cdot|\x)\right)\leq\max_{\y,\y'\in \mathcal{Y}}\{\Vert \y- \y'\Vert_2\}\Vert\hat{\pi}(\cdot|\x)-\pi(\cdot|\x)\Vert_1 =  \eta \Vert\hat{\pi}(\cdot|\x)-\pi(\cdot|\x)\Vert_1
\end{equation}
for any $\x$, where we denote $\eta=\max_{\y,\y'\in \mathcal{Y}}\{\Vert \y- \y'\Vert_2\}$. Therefore,
\begin{equation}
\begin{split}
    \E_{\x\sim p}W_2\left(\hat{\pi}(\cdot|\x),\pi(\cdot|\x)\right)&\leq \eta \E_{\x\sim p}\Vert\hat{\pi}(\cdot|\x)-\pi(\cdot|\x)\Vert_1\\
    &= \eta \int p(\x)\int\big|\hat{\pi}(\y|\x)-\pi(\y|\x)\big|\dif \y\dif \x \\
    &= \eta\int \big|p(\x)\hat{\pi}(\y|\x)- p(\x)\pi(\y|\x)\big|\dif \y\dif\x\\
    & = \eta\int \big|\hat{\pi}(\x,\y)-\pi(\x,\y)\big|\dif\x\dif\y\\
    &= \eta\big\Vert \hat{\pi}-\pi\big\Vert_1.
    \end{split}
\end{equation}
By virtue to Theorem 4.3 in~\cite{daniels2021score}, we have
\begin{equation}
    \Vert \hat{\pi}-\pi\Vert_1 \leq \frac{1}{\kappa}\big\Vert \nabla_{\hat{\pi}}\mathcal{L}(\hat{\pi},{u}_{\hat{\omega}},{v}_{\hat{\omega}})\big\Vert_1.
\end{equation}
We therefore  have 
\begin{equation}
    \E_{\x\sim p}W_2(\hat{\pi}(\cdot|\x),\pi(\cdot|\x))\leq \frac{\eta}{\kappa}\big \Vert \nabla_{\hat{\pi}}\mathcal{L}(\hat{\pi},{u}_{\hat{\omega}},{v}_{\hat{\omega}})\big\Vert_1.
\end{equation}
Let $C_1 = \frac{\eta}{\kappa}$, we have 
\begin{equation}\label{eq_a:bound_c1}
    \E_{\x\sim p}W_2(\hat{\pi}(\cdot|\x),\pi(\cdot|\x))\leq C_1\big \Vert \nabla_{\hat{\pi}}\mathcal{L}(\hat{\pi},{u}_{\hat{\omega}},{v}_{\hat{\omega}})\big\Vert_1.
\end{equation}

Combining Eqs.~\eqref{eq_a:triangle}, \eqref{eq_a:bound_c2_c3}, and~\eqref{eq_a:bound_c1}, we have
\begin{equation}
    \begin{split}
    \E_{\x\sim p}W_2\left(p^{\rm sde}(\cdot|\x),\pi(\cdot|\x)\right)
 \leq & C_1 \big\Vert \nabla_{\hat{\pi}}\mathcal{L} (\hat{\pi}, u_{\hat{\omega}}, v_{\hat{\omega}})\big\Vert_1+ \sqrt{C_2 \mathcal{J}_{\rm CSM}(\hat{\theta})} \\
  +& C_3\E_{\x\sim p}W_2\left(p_T(\cdot|\x),p_{\rm prior}\right).
\end{split}
\end{equation}
The proof is completed.

\section{Experimental Details}
We provide the details for learning the potentials $u_{\omega}(\x), v_{\omega}(\y)$, training the conditional score-based model $s_{\theta}(\y;\x,t)$, generating data in inference, and computing the metric Acc. All the experiments are conducted using 2 NVIDIA Tesla V100 32GB GPUs. The codes are in pytorch~\cite{paszke2019pytorch}.
\subsection{Details for Toy Data Experiment in Figure 2}
\textbf{Architectures of $u_{\omega}, v_{\omega}$.} The architectures of  both of $u_{\omega}$ and $ v_{\omega}$ are FC(1,1024) $\rightarrow$ Tanh $\rightarrow$ FC(1024,1), where FC($a,b$) is the fully-connected layer with input/output dimension of $a/b$ and Tanh is the activation function. 

\textbf{Details for learning $u_{\omega}, v_{\omega}$.} We use the $L_2$-regularized unsupervised OT where $c$ is taken as the squared $L_2$-distance. The learning rate is 1e-5. The batch size $B'$ is set to 256. The Adam algorithm is employed to update the parameters. 

\textbf{Architecture of $s_{\theta}$.} The backbone of $s_{\theta}$ is FC(1,512) $\rightarrow$ SiLU $\rightarrow$ FC(512,512) $\rightarrow$ SiLU $\rightarrow$ FC(512,1), where SiLU is the activation function. We add the embedding of time $t$ and condition $\x$ to the activation of SiLU. The embedding block for $t$ is GaussianFourierProjection(256) $\rightarrow$ FC(256,512) $\rightarrow$ SiLU $\rightarrow$ FC(512,512). The embedding block for $\x$ is FC(1,512) $\rightarrow$ SiLU $\rightarrow$ FC(512,512) $\rightarrow$ SiLU $\rightarrow$ FC(512,512). The GaussianFourierProjection has been adopted in~\cite{song2020score}. 

\textbf{Details for training $s_{\theta}$ and inference.} We take the VE-SDE with $\alpha=25$, and $T=1$. We set $w_t=\sigma_t^2$, the batch size $B=32$, $L=10B$ in Algorithm~\ref{alg:alg1}. The learning rate is 1e-4. The Adam algorithm and the exponential moving average for model parameters with decay=0.999 are applied. We take the Euler-Maruyama method to perform the reverse SDE for generating data in inference. The initial state $\y_T$ is sampled from the $p_{\rm prior}=\mathcal{N}(0,\sigma_T^2 \mathbf{I})$.

\subsection{Details for Unpaired Super-Resolution}
\textbf{Architectures of $u_{\omega}, v_{\omega}$.} The architectures of $u_{\omega}$ and $ v_{\omega}$ are  FC(12288,512) $\rightarrow$ SiLU $\rightarrow$ FC(512,512) $\rightarrow$ SiLU $\rightarrow$ FC(512,512) $\rightarrow$ SiLU $\rightarrow$ FC(512,512) $\rightarrow$ SiLU $\rightarrow$ FC(512,512) $\rightarrow$ SiLU $\rightarrow$ FC(512,512) $\rightarrow$ SiLU $\rightarrow$ FC(512,1). We reshape the input images from size (64,64,3) to size 12288. The design of the architectures is inspired by~\cite{daniels2021score}.

\textbf{Details for learning $u_{\omega}, v_{\omega}$.} We use the $L_2$-regularized unsupervised OT where $c$ is taken as the mean squared $L_2$-distance (following~\cite{daniels2021score}), and $\epsilon$ is set to 1e-7. The learning rate is 1e-6. The batch size $B'$ is set to 64. The Adam algorithm is employed to update the parameters. 

\textbf{Architecture of $s_{\theta}$.} The backbone of $s_{\theta}$ is based on the architecture of DDIM~\cite{song2020denoising} on CelebA dataset for unconditional image generation. We apply the condition to the backbone by concatenating the degenerated image $\x$ with the noisy image $\y_t$ as input, inspired by~\cite{saharia2022image} that tackles the paired super-resolution.


\textbf{Details for training $s_{\theta}$ and inference.} We take the VP-SDE with $\beta_{\rm min}=0.1, \beta_{\rm max}=20$, and $T=1$. We set $w_t=\sigma_t^2$, the batch size $B=64$ in Algorithm~\ref{alg:alg2}. The learning rate is $2e-4$. The Adam algorithm and the exponential moving average for model parameters with decay=0.999 are applied. 
To facilitate the training, we take the trained model in~\cite{song2020denoising} on CelebA images as initialization.
In inference, we take the sampling method in DDIM to perform the reverse SDE to generate data. Following~\cite{zhao2022egsde,meng2021sdedit}, we add noise to the low-resolution images by sampling $\y_M$ from $p_{M|0}(\y_M|\x)$ as the initial state. $M$ is set to 0.2.

\subsection{Details for Semi-paired Image-to-Image Translation on Animal Images}
In experiments, we randomly choose 1000/150 images for each species for training/testing.

\textbf{Architectures of $u_{\omega}, v_{\omega}$.} The architectures of $u_{\omega}$ and $ v_{\omega}$ consist of a feature extractor and a head. We take the image encoder ``ViT-B/32'' of CLIP~\cite{radford2021learning} as the feature extractor. The feature extractor is fixed in training.  The architecture of the head is the same as that of $u_{\omega}, v_{\omega}$ for unpaired super-resolution except that the input dimension is 512.

\textbf{Details for learning $u_{\omega}, v_{\omega}$.} We use the $L_2$-regularized semi-supervised OT where $c$ is taken as the cosine distance of extracted features by the above feature extractor, and $\epsilon$ is set to 1e-5. The learning rate is 1e-6. The batch size $B'$ is set to 64. The Adam algorithm is employed to update the parameters. 

\textbf{Architecture of $s_{\theta}$.} The architecture of $s_{\theta}$ is based on the architecture of model of ILVR~\cite{choi2021ilvr} on dog images for unconditional image generation. We add the embedding of condition $\x$ to the output of each residual block. The embedding block for condition $\x$ comprises the feature extractor as mentioned above followed by an embedding module. The architecture of the embedding module is FC(512,512) $\rightarrow$ SiLU $\rightarrow$ FC(512,512).

\textbf{Details for training $s_{\theta}$, inference, and computing the Acc.} We take the VP-SDE with $\beta_{\rm min}=0.1, \beta_{\rm max}=20$, and $T=1$. We set $w_t=\sigma_t^2$, the batch size $B=16$ in Algorithm~\ref{alg:alg2}. The learning rate is $2e-5$. The Adam algorithm and the exponential moving average for model parameters with decay=0.999 are applied. To facilitate the training, we take the trained model in~\cite{choi2021ilvr} on dog images as initialization.
In inference, we take the sampling method in DDIM to perform the reverse SDE to generate data. The initial state $\y_T$ is sampled from the $p_{\rm prior}=\mathcal{N}(0,\mathbf{I})$. To compute the metric Acc, we classify the translated images using CLIP (``ViT-B/32'') into the candidate classes of lion, tiger, and wolf. We then compute the precision against the ground-truth translated classes.

\subsection{Details for Semi-paired Image-to-Image Translation on Digits}
\textbf{Architectures of $u_{\omega}, v_{\omega}$.} The architectures of $u_{\omega}$ and $ v_{\omega}$ are the same as the architectures of $u_{\omega}, v_{\omega}$ for unpaired super-resolution except that the input dimension is 784. We reshape the input images from size (28,28) to size 784.

\textbf{Details for learning $u_{\omega}, v_{\omega}$.} We use the $L_2$-regularized semi-supervised OT where $c$ is taken as the cosine distance of extracted features by a pre-trained feature extractor, and $\epsilon$ is set to 1e-5. 
The learning rate is 1e-6. The batch size $B'$ is set to 64. The Adam algorithm is employed to update the parameters. We train auto-encoders (consisting of an encoder and a decoder) for MNIST and Chinese-MNIST respectively, and the encoder is taken as the feature extractor. The architecture of the encoder is Conv(1,64,4,2,0) $\rightarrow$ BN $\rightarrow$ SiLU $\rightarrow$ Conv(64,128,4,2,0) $\rightarrow$ BN $\rightarrow$ SiLU $\rightarrow$ Conv(128, 128,3,1,0)  $\rightarrow$ BN  $\rightarrow$ SiLU, where ``BN'' is the Batch Normalization layer.
The architecture of the encoder is Conv(128,128,3,1,0) $\rightarrow$ BN $\rightarrow$ SiLU $\rightarrow$ Tconv(128,64,4,2,0) $\rightarrow$ BN $\rightarrow$ SiLU $\rightarrow$ Tconv(64,1,4,2,0)  $\rightarrow$ Sigmoid, where Tconv is the transposed convolutional layer, and Sigmoid is the activation function. We use Adam algorithm to train the auto-encoder with learning rate 1e-4.

\textbf{Architecture of $s_{\theta}$.} The backbone of $s_{\theta}$ is the architecture of model~\cite{song2020score} on MNIST for unconditional image generation. We add the embedding of condition $\x$ to the output of each residual block. The embedding block for condition $\x$ is FC(784,512) $\rightarrow$ SiLU $\rightarrow$ FC(512,512) $\rightarrow$ SiLU $\rightarrow$ FC(512,256).

\textbf{Details for training $s_{\theta}$, inference, and computing the Acc.} We take the VE-SDE with $\alpha=25$, and $T=1$. We set $w_t=\sigma_t^2$, the batch size $B=32$ in Algorithm~\ref{alg:alg2}. The learning rate is 1e-4. The Adam algorithm and the exponential moving average for model parameters with decay=0.999 are applied. In inference, we take the Predictor-Corrector algorithm in~\cite{song2020score}  to perform the reverse SDE to generate data, where the predictor is taken as the Euler-Maruyama method. The initial state $\y_T$ is sampled from the $p_{\rm prior}=\mathcal{N}(0,\sigma_T^2 \mathbf{I})$. To compute the metric Acc, we classify the translated images using a classifier (LeNet) trained on Chinese-MNIST. We then compute the precision against the ground-truth translated classes.

\section{Additional Experimental Analysis and Results}
\subsection{Additional Experimental Analysis}
\begin{figure}[H]
    \centering
    \includegraphics[width=0.9\columnwidth]{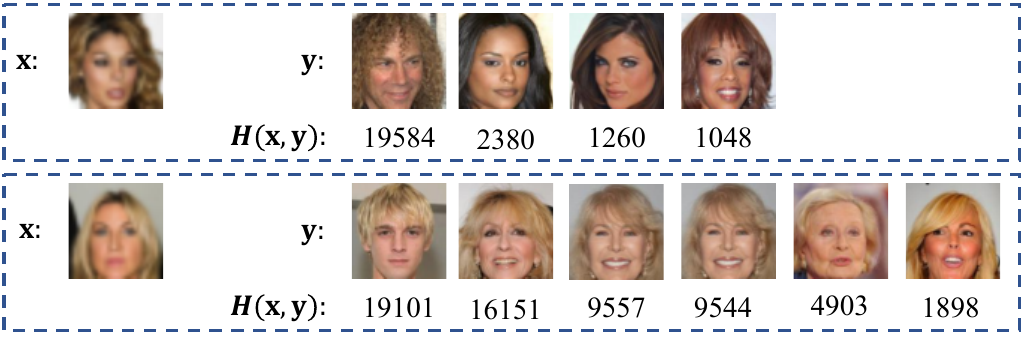} 
    \caption{Examples of guided high-resolution images (\ie, $H(\x,\y)>0$) chosen from B1 based on OT for the given degenerated low-resolution image $\x$ from A0 in training. Note that considering the numerical issue, we choose the guided high-resolution images $\y$ such that $H(\x,\y)>0.01$.
    }
    \label{fig_a:guide_image}
\end{figure}

\paragraph{Guided images sampled based on OT.} We show the examples of guided high-resolution images sampled based on OT in Fig.~\ref{fig_a:guide_image}. We can observe that the guided high-resolution images share similar structures to the given degenerated image.

\paragraph{What happens to the compatibility on the source data with no good target to be paired?}
To figure out what happens to the compatibility function $H$ when there is no good target data, we conduct the following experiments. Firstly, we count the number of source samples satisfying that there is no target sample such that $H>0.001$, in CelebA dataset. We find that 25.9\% of source samples meets such condition. Note that the other source samples are often with $H$ larger than 1000 on some target samples (since the $\epsilon$ is 1e-7 in Eq. (7)). Secondly, we add noisy images to the source dataset to train the potentials $u_{\omega}$ and $v_{\omega}$, and count the ratio of noisy images satisfying $H<0.001$ on all target samples. The results are reported in Table~\ref{tab_a:compatibility_on_noise}. The noisy images are generated from the standard normal distribution and with the same shape as the source images.

\begin{table}[H]
    \centering
    \caption{Ratio of noisy images with $H<0.001$ when adding varying numbers of noisy images to the source dataset.}
    \setlength{\tabcolsep}{4.5pt}
    \begin{tabular}{l|ccccc}
    \toprule
         Number of noisy images : Number of clean images&0.1 : 1&0.2 : 1&0.3 : 1&0.4 : 1&0.5 : 1\\
    \midrule
         Ratio of noisy images assigned with $H<0.001$&89.3\%&85.6\%&83.9\%&81.6\%&80.2\%\\
    \bottomrule
    \end{tabular}
    
    \label{tab_a:compatibility_on_noise}
\end{table}

It can be seen that more than 80\% of noisy images that are with no good target data are assigned with near-to-zero $H$ ($H<0.001$), when the ratio of numbers of noisy images to clean images is in [0.1,0.5]. 

\paragraph{Empirical comparison of the ``soft'' and ``hard'' coupling relationship.}
To study how sparse $H$ is, for each target image $\mathbf{y}$, we denote the number of source image $\mathbf{x}$ with "non-zero $H$" as $n_{\mathbf{y}}$ (i.e., $n_{\mathbf{y}}=|\{\mathbf{x}:H(\mathbf{x},\mathbf{y})>0.001\}|$, considering numerical issues) in CelebA dataset. The histogram of $n_{\mathbf{y}}$ is shown in the Table~\ref{tab_a:soft_hard}.

\begin{table}[H]
    \centering
    \caption{Histogram of number $n_{\mathbf{y}}$ of source images with "non-zero $H$" for target image $\mathbf{y}$, where the total numbers of both source images and target images are 80k.}
    \setlength{\tabcolsep}{4.5pt}
    \begin{tabular}{l|cccccc}
    \toprule
         Bins for $n_{\mathbf{y}}$&[0,10)&[10,20)&[20,50)&[50,100)&[100,600)&[600,80k]\\
    \midrule
         Frequency&59600&8064&8468&3537&1716&0\\
    \bottomrule
    \end{tabular}
    
    \label{tab_a:soft_hard}
\end{table}

We can see from Table~\ref{tab_a:soft_hard} that all the target images are with $n_{\mathbf{y}}\leq 600$, and more than 70\% of target images are with  $n_{\mathbf{y}}\leq 10$. This implies that for each $\mathbf{y}$ in more than 70\% target images, there are no more than 10 among 80K source images $\mathbf{x}$ satisfying $H(\mathbf{x},\mathbf{y})>0.001$. So $H$ is sparse to some extent. We also count the number of target images with $n_{\mathbf{y}}=1$ ($n_{\mathbf{y}}=1$ means that each target image is paired with one source image), which is 8579 (around 10\%). These empirical results indicate that $H$ may provide a ''soft" coupling relationship, since there may exist multiple source images with "non-zero $H$" for most target images. 

\begin{figure}[H]
    \centering
    \subfigure[$lr=1e-5$]{\includegraphics[width=0.3\columnwidth]{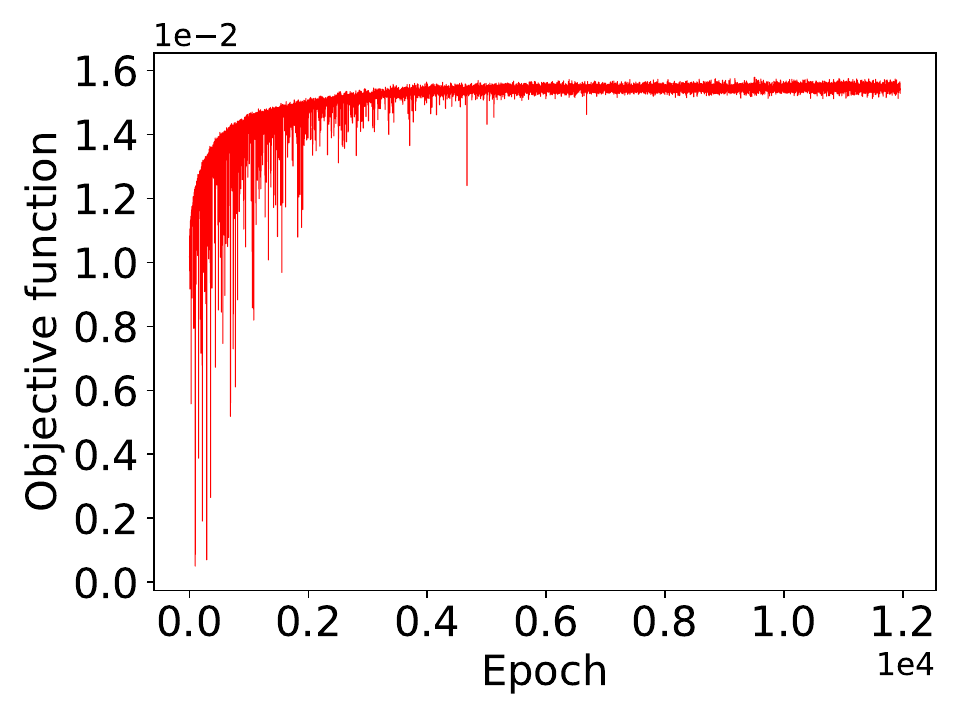} \label{fig_a:lr_1e-5}}
    \subfigure[$lr=1e-6$]{\includegraphics[width=0.3\columnwidth]{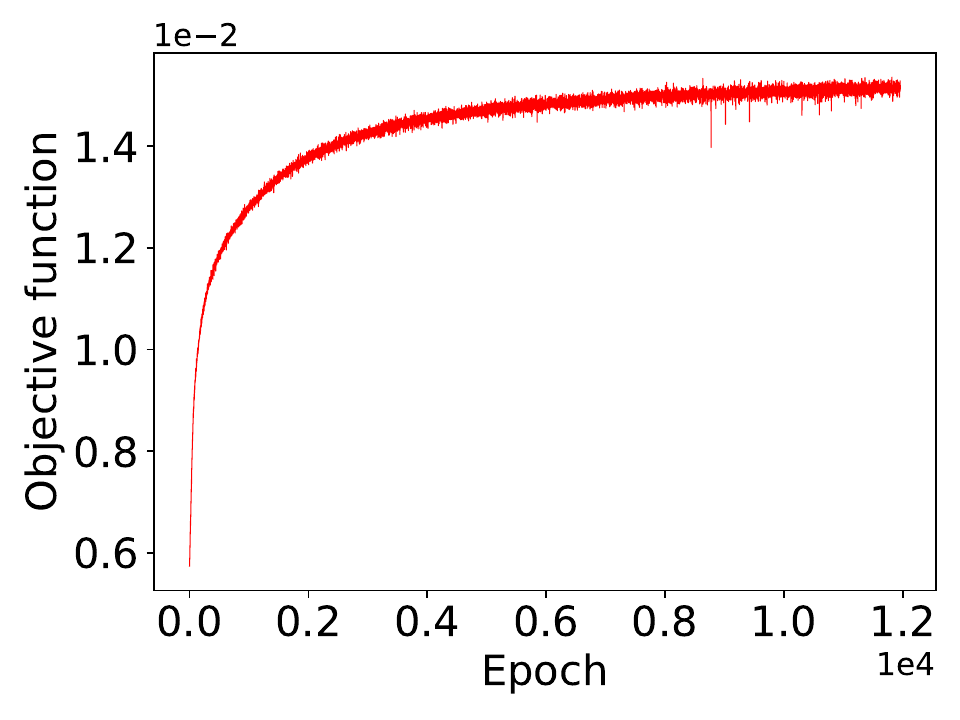} \label{fig_a:lr_1e-6}}
    \subfigure[Relative difference of $H$]{\includegraphics[width=0.3\columnwidth]{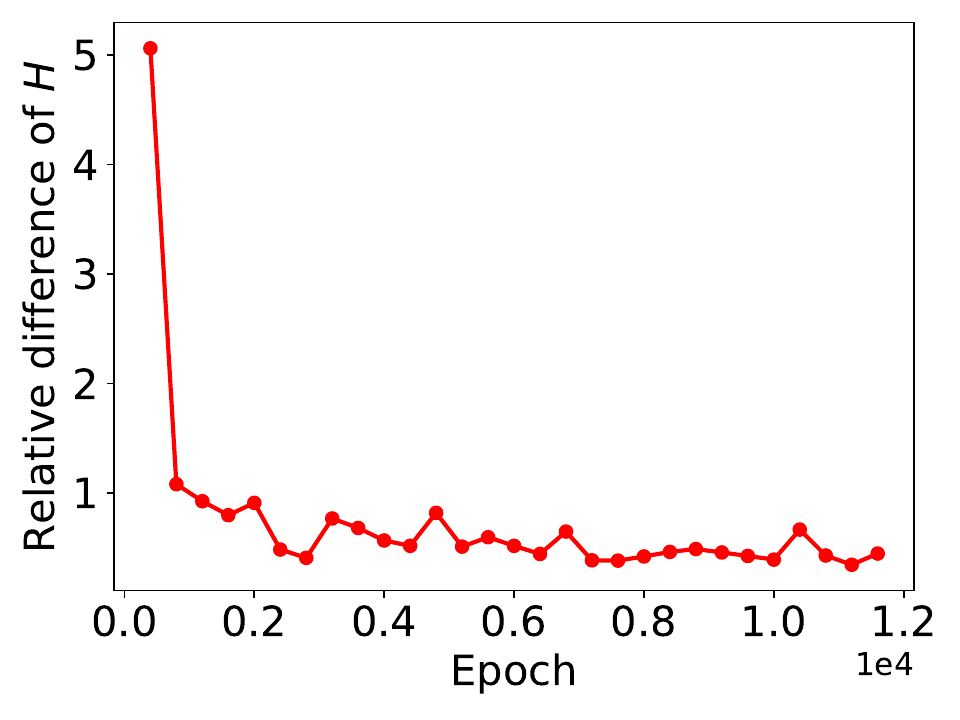} \label{fig_a:H_diff}}
    \caption{ (a-b) Curves of objective function in Eq.~(6) under learning rates $lr=1e-5$ and $lr = 1e-6$ with $\epsilon=1e-5$. (c) Relative difference of $H$ in training. The relative difference of $H$ is defined as $\frac{\|H^i-H^{i+\Delta}\|_{p,q}}{\|H^i\|_{p,q}}$, where the norm $\|H\|_{p,q}=\left[\mathbb{E}_{p}\mathbb{E}_qH(\x,\y)^2\right]^{1/2}$ (\ie, the $L_2$-norm of functions on the sample space associated to measure $p\otimes q$). $H^i$ is the function at training step $i$. To reduce the computational cost, we set $\Delta=10000$. $p$ and $q$ are distributions of source and target training data. 
    }
    \label{fig_a:convergence}
\end{figure}

\paragraph{Stability and convergence of training process for learning $u_{\omega}$ and $v_{\omega}$.}
We show the objective function (Eq. (6)) in training in Figs.~\ref{fig_a:convergence}(a-b). We can see that the objective function first increases and then converges, under learning rates 1e-5 and 1e-6. We notice that different $u_{\omega}$ and $v_{\omega}$ may yield the same $H$, (\eg, $u_{\omega}(\mathbf{x})+c$ and  $v_{\omega}(\mathbf{y})-c$ yield the same $H(\mathbf{x},\mathbf{y})$ as $u_{\omega}(\mathbf{x})$ and $v_{\omega}(\mathbf{y})$, as in Eq. (7)). We then show the relative change of $H$ in training in Fig.~\ref{fig_a:H_diff}. We can see that the relative difference of $H$ first decreases and fluctuates near to zero, which may be because the optimization is based on approximated gradients over mini-batch. The $\frac{1}{\epsilon}$ ($\epsilon=$ 1e-5 or 1e-7 in experiments) in Eq. (6) may yield large gradients. We then choose a small learning rate to stabilize the training. 

\begin{figure}[H]
    \centering
    \subfigure[$\epsilon=1e-5$]{\includegraphics[width=0.3\columnwidth]{rebuttal_figures/lr_5e-07.pdf} \label{fig_a:reg_1e-5}}
    \subfigure[$\epsilon=1e-6$]{\includegraphics[width=0.3\columnwidth]{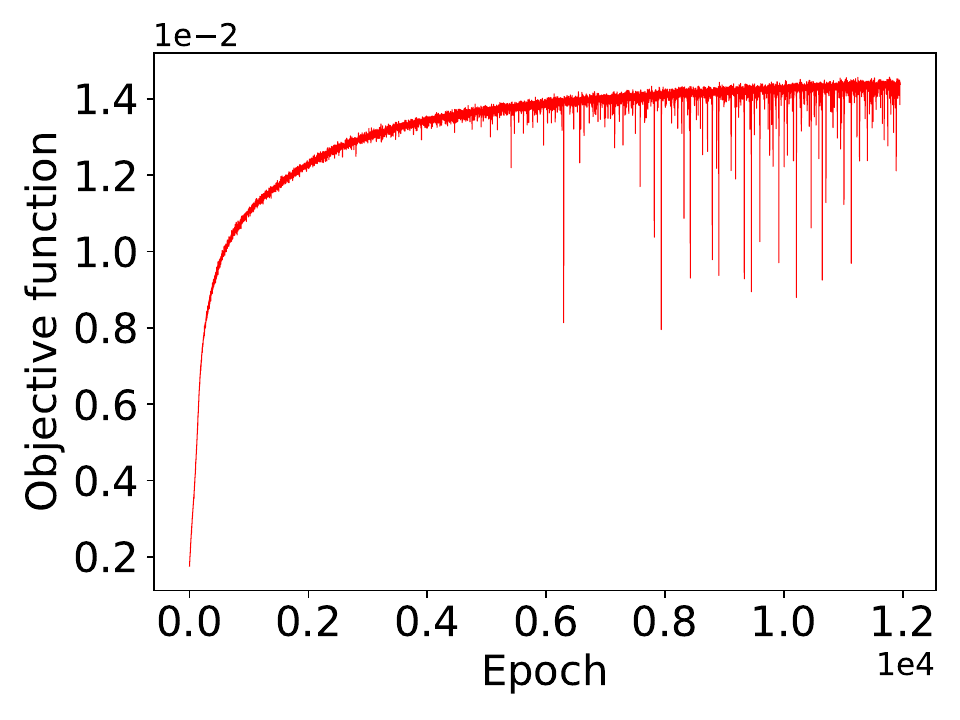} \label{fig_a:reg_1e-6}}
    \subfigure[$\epsilon=1e-7$]{\includegraphics[width=0.3\columnwidth]{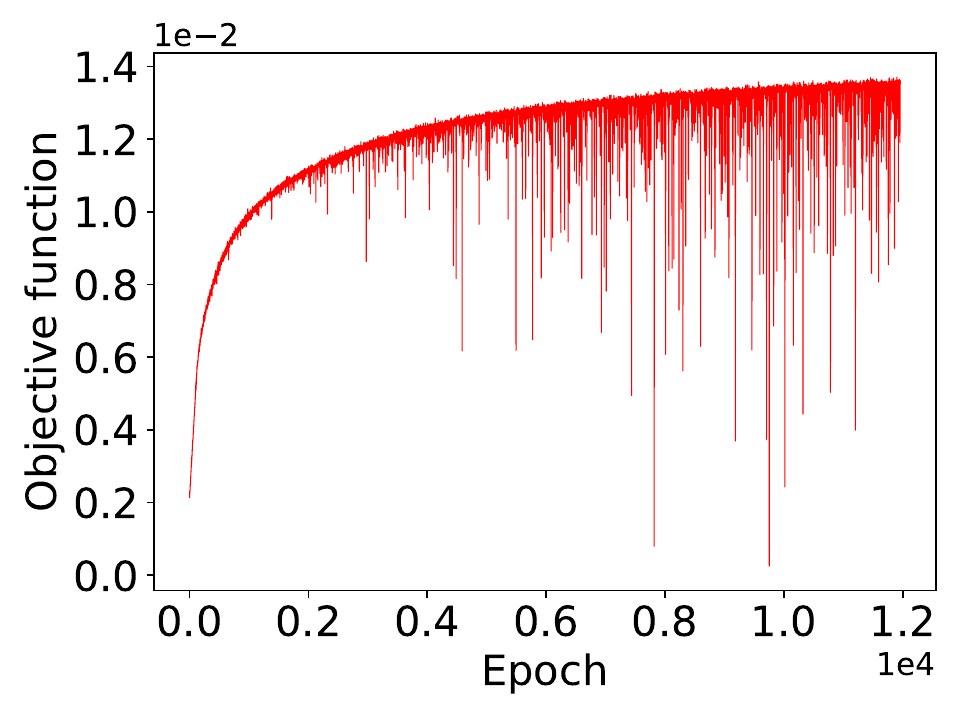} \label{fig_a:reg_1e-7}}
    \caption{The curves of objective function in training under varying $\epsilon$ with learning rate $lr=1e-6$. 
    }
    \label{fig_a:curves_to_epsilon}
\end{figure}

\paragraph{On the choice of $\epsilon$.}
To better approach the original OT in Eqs. (2-3) by the $L_2$-regularized OT in Eq. (5) so that the OT guidance could be better achieved, the $\epsilon$ should be small. However, due to the term $\frac{1}{\epsilon}$ in the objective function in Eq. (6), smaller $\epsilon$ may suffer from numerical issues in training. As a balance, we empirically choose a $\epsilon$ from candidate values {1e-5, 1e-6, 1e-7} such that the training is more stable.  We show the objective function curves under varying $\epsilon$ in Fig.~\ref{fig_a:curves_to_epsilon}. The training curves seem to be stable in general. We have also reported the results with varying $\epsilon$ in Table~\ref{tab_a:ablation_epsilon_animal}. From Table~\ref{tab_a:ablation_epsilon_animal}, we can see that FID ranges in [13.68, 14.56] (which seems to be stable) for $\epsilon$ in [1e-7,1e-3]. We can also see that Acc is similar for $\epsilon$ in {1e-7, 1e-6,1e-5}, and decreases as $\epsilon$ increases from 1e-5 to 1e-3. 

\begin{table}[H]
    \centering
    \caption{Results of OTCS using varying $\epsilon$.}
    \begin{tabular}{c|ccccc}
    \toprule
         $\epsilon$& 1e-7&1e-6&1e-5&1e-4&1e-3\\
    \midrule
         FID $\downarrow$&14.56&14.12&13.68&13.52&13.91  \\
         Acc $\uparrow$&95.11&96.00&96.44 &90.22&77.78\\
    \bottomrule
    \end{tabular}
    
    \label{tab_a:ablation_epsilon_animal}
\end{table}

\paragraph{Computational cost.}
We report the computational time cost of our training process in this paragraph. As illustrated in Algorithm 2 in the Appendix~\ref{sec:app_A}, our method consists of three processes in training: (1) \textit{learning the potentials $u_{\omega}$ \& $v_{\omega}$}, (2) \textit{computing $H$ \& storing the target sample indexes with non-zero $H (H>0.001)$ for each source sample}, and (3) \textit{training the score-based model $s_{\theta}$}. We report the computational time cost of these three processes in the following Table~\ref{tab_a:computational_cost}. 

\begin{table}[H]
    \centering
    \caption{Computational time cost of training processes.}
    \begin{tabular}{c|ccc}
    \toprule
         Dataset&Learning $u_{\omega}$ \& $v_{\omega}$ (30w steps)&Computing \& storing $H$&Training $s_{\theta}$ (60w steps)\\
    \midrule
        CelebA&3.5 hours&0.5 hours&5 days\\
        Animal&2.0 hours&0.05 hours&5 days\\
    \bottomrule
    \end{tabular}
    
    \label{tab_a:computational_cost}
\end{table}

From Table~\ref{tab_a:computational_cost}, we can see that (1) learning $u_{\omega}$ \& $v_{\omega}$ and (2) computing \& storing $H$ takes no more than 4 hours. Similarly to the other diffusion approaches, (3) training our score-based model  $s_{\theta}$ takes a few days. \\
\textit{\textbf{Computational time of each operation in a single step of training $s_{\theta}$}}. In each step of training the score-based model $s_{\theta}$, we sequentially (1) sampling the index of target sample with probability proportional to $H$ for a randomly selected source sample index (\textit{sampling index}), then (2) load corresponding images (\textit{loading image}), and (3) finally feed data to network and update model parameters (\textit{updating network}). Compared with the training of score-based model for paired setting, our training additionally contains the operation of sampling index. From Table~\ref{tab_a:computational_cost1}, we can see that sampling index takes much less time than updating network. 

\begin{table}[H]
    \centering
    \caption{Computational time of operations in a single step of training $s_{\theta}$ on Animal dataset.}
    \begin{tabular}{ccc}
    \toprule
         Sampling index&Loading image&Updating network\\
    \midrule
        0.0005 seconds&0.01 seconds&0.7 seconds\\
    \bottomrule
    \end{tabular}
    
    \label{tab_a:computational_cost1}
\end{table}

\subsection{Additional Results on Toy Data Experiments}
\begin{figure}[H]
    \centering
    \subfigure[$\epsilon=0.1$]{\includegraphics[width=0.3\columnwidth]{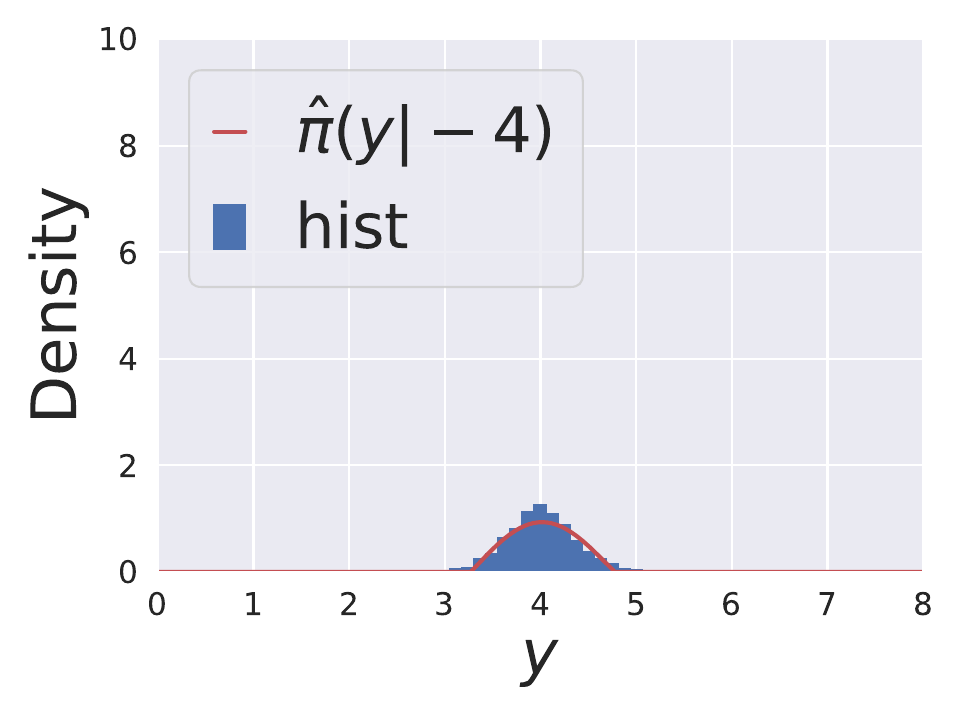} \label{fig_a:otcs_0.1}}
    \subfigure[$\epsilon=0.01$]{\includegraphics[width=0.3\columnwidth]{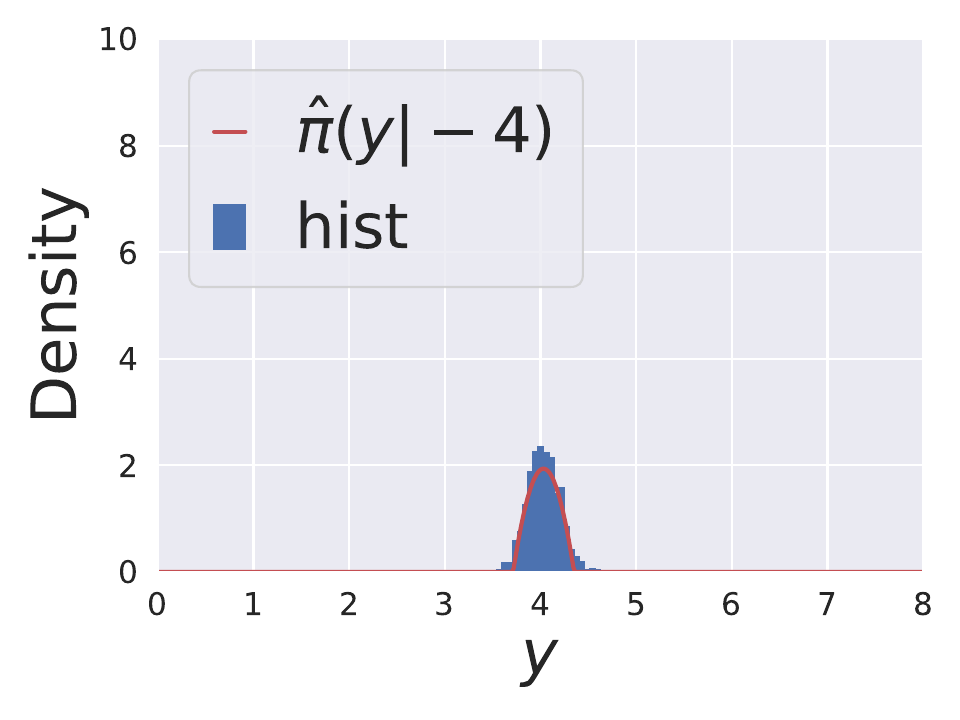} \label{fig_a:otcs_0.01}}
    \subfigure[$\epsilon=0.001$]{\includegraphics[width=0.3\columnwidth]{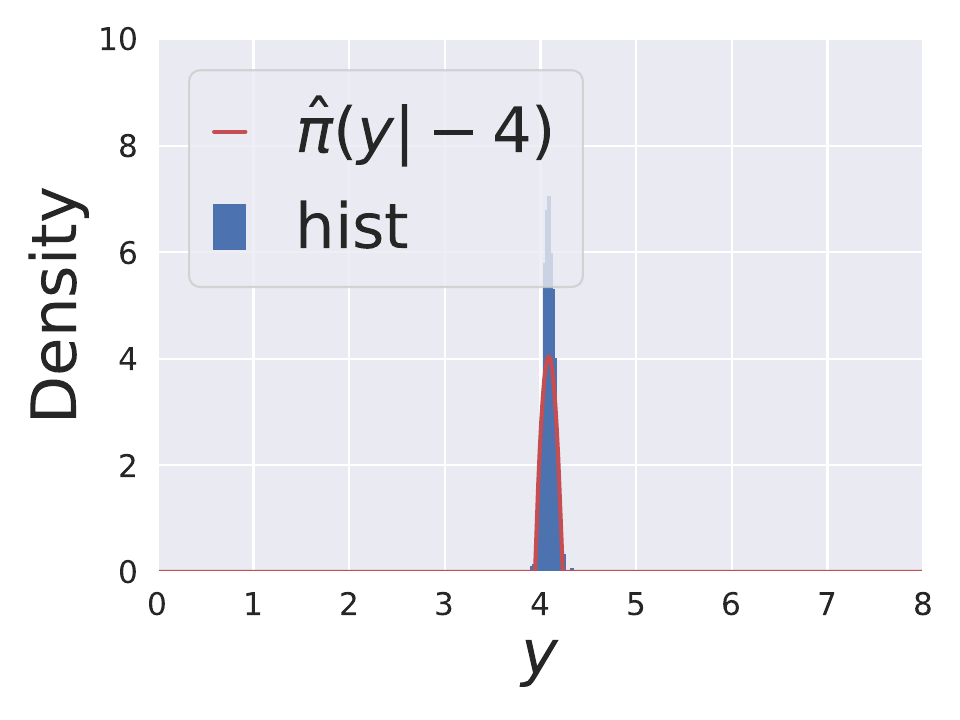} \label{fig_a:otcs_0.001}}
    \caption{The histogram (``hist'') of generated samples by our proposed OTCS and the estimated conditional transport plan $\hat{\pi}(\y|-4)$ under varying $\epsilon$.
    }
    \label{fig_a:1d_example_epsilon}
\end{figure}

\textbf{Results of OTCS under varying $\epsilon$.} We show the results of OTCS under varying $\epsilon$ in Fig.~\ref{fig_a:1d_example_epsilon}. We can see that the histogram of generated samples by OTCS fits the estimated conditional transport plan when $\epsilon$ is 0.01, 0.001, and 0.0001.

\begin{figure}[H]
    \centering
    \subfigure[$\x=-5$]{\includegraphics[width=0.3\columnwidth]{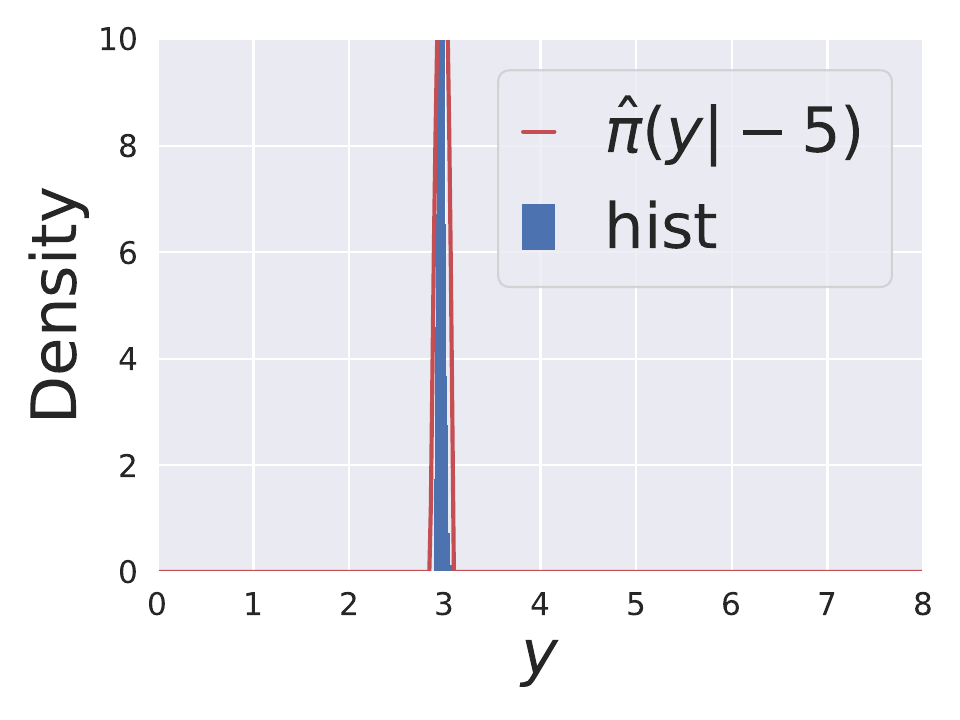} \label{fig_a:otcs_-5}}
    \subfigure[$\x =-4$]{\includegraphics[width=0.3\columnwidth]{figures/OTCS_0.0001.pdf} \label{fig_a:otcs_-4}}
    \subfigure[$\x=-3$]{\includegraphics[width=0.3\columnwidth]{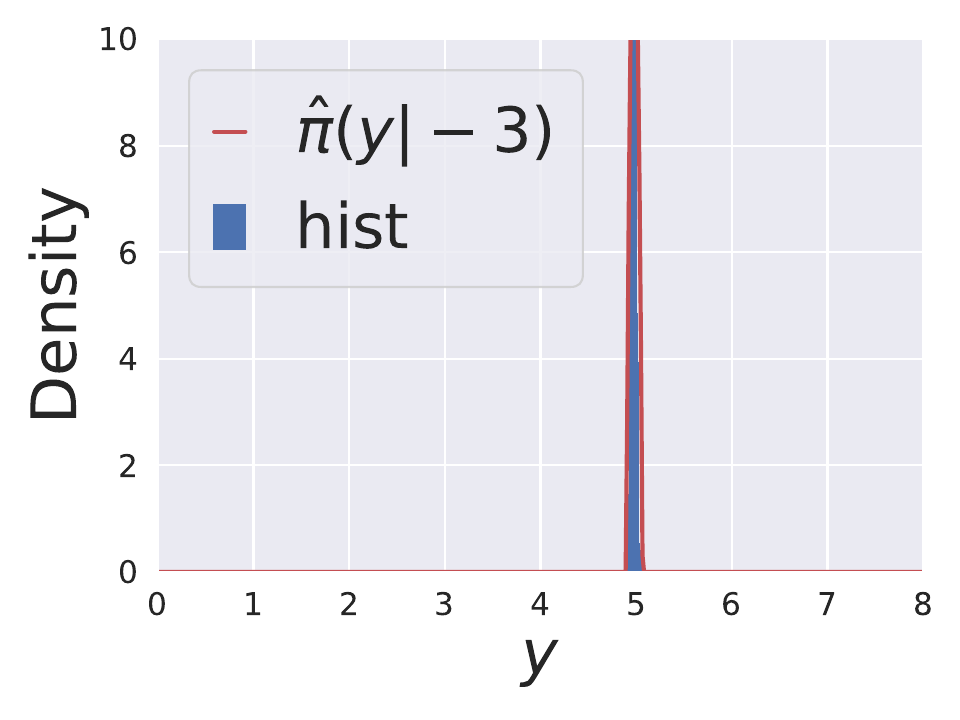} \label{fig_a:otcs_-3}}
    \caption{The histogram (``hist'') of generated samples by our proposed OTCS and the estimated conditional transport plan $\hat{\pi}(\y|\x)$ under varying condition $\x$. $\epsilon=0.0001$ in this experiment.
    }
    \label{fig_a:1d_example_condition}
\end{figure}

\textbf{Results of OTCS under varying conditions.} We show the results of OTCS for varying condition $\x$ in Fig.~\ref{fig_a:1d_example_condition}. We can see that the histogram of generated samples by OTCS fits the estimated conditional transport plan for $\x=-5, -4, -3$.

\subsection{Additional Results in Unpaired Super-Resolution}
\begin{figure}[H]
    \centering
    \includegraphics[width=1.0\columnwidth]
    {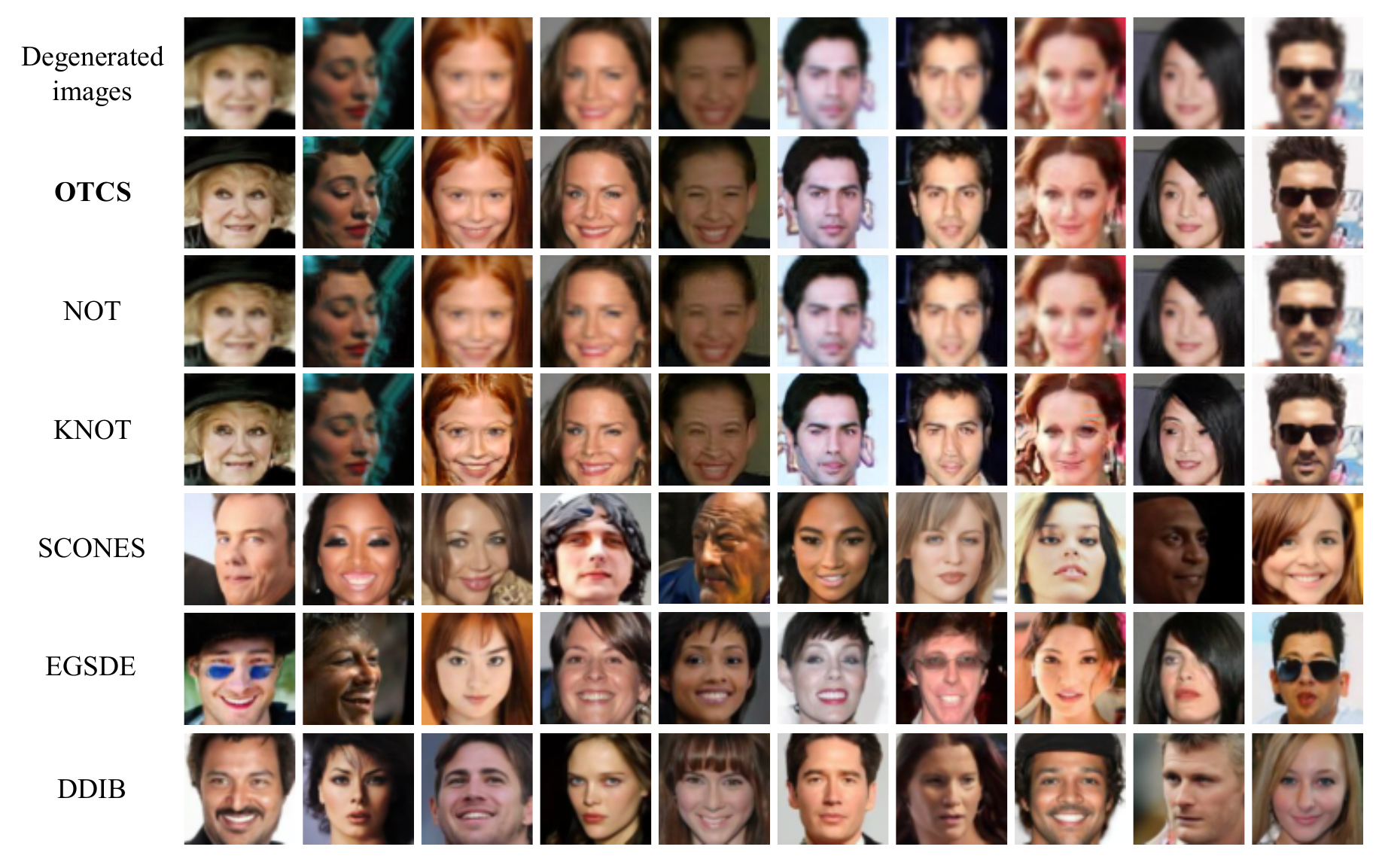}
    \caption{Translated images by our proposed OTCS, adversarial training-based OT methods of NOT and KNOT, and diffusion-based methods of SCONES, EGSDE, and DDIB.
    }
    \label{fig_a:images_celeba_method}
\end{figure}

\begin{figure}[H]
    \centering
    \includegraphics[width=1.0\columnwidth]
    {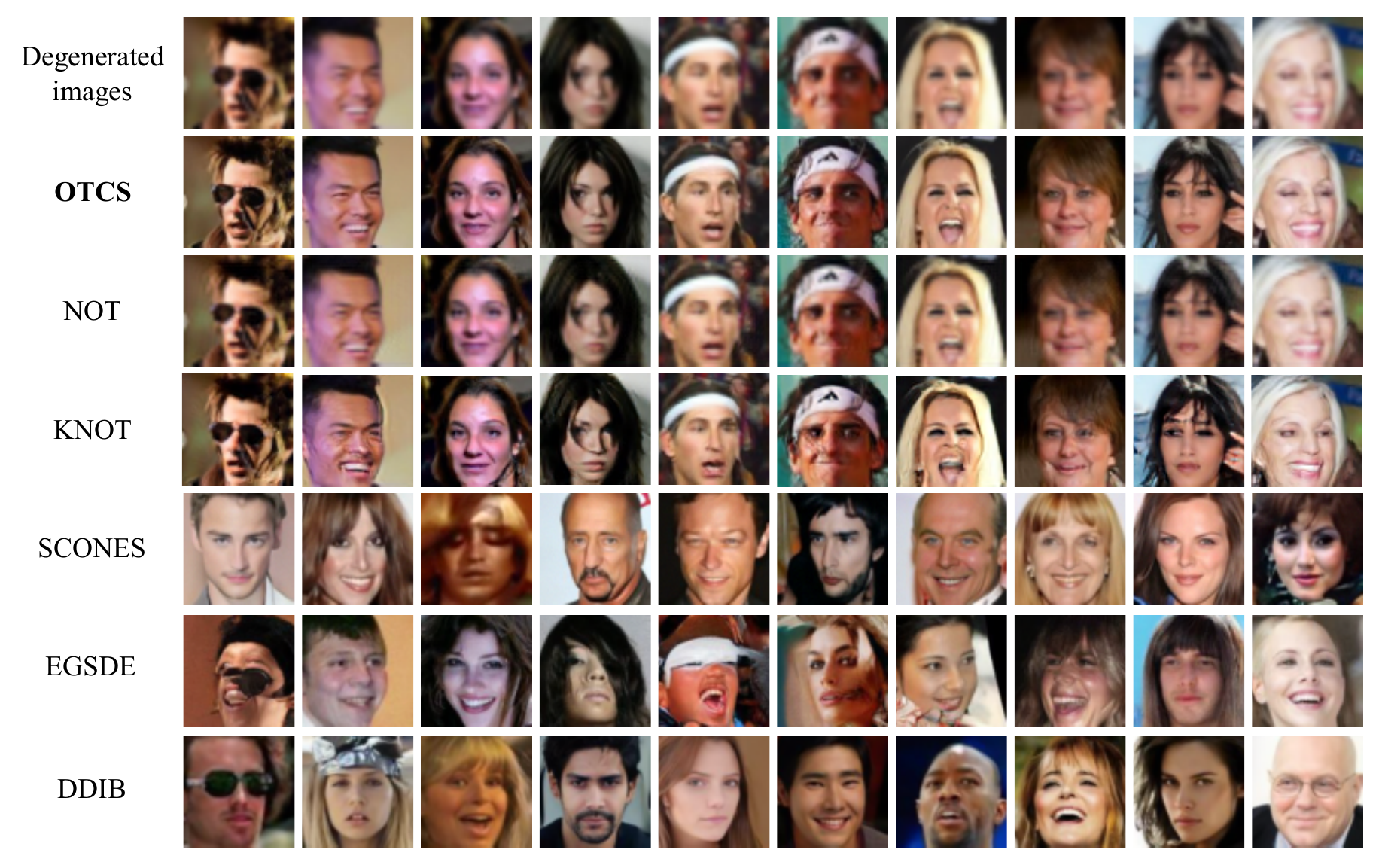}
    \caption{Translated images by our proposed OTCS, adversarial training-based OT methods of NOT and KNOT, and diffusion-based methods of SCONES, EGSDE, and DDIB.
    }
    \label{fig_a:images_celeba_method1}
\end{figure}

\paragraph{Results of different methods in unpaired super-resolution.} In Figs.~\ref{fig_a:images_celeba_method} and~\ref{fig_a:images_celeba_method1}, we visualize the translated images by our proposed OTCS,  adversarial training-based OT methods of NOT and KNOT, and diffusion-based methods of SCONES, EGSDE, and DDIB. We can see that OTCS, NOT, and KNOT better preserve the identity/structure than SCONES, EGSDE, and DDIB. OTCS produces clearer translated images than NOT. The translated images by KNOT have artifacts (please zoom in on the figure to see the artifacts).

\begin{figure}[t]
    \centering
    \includegraphics[width=0.99\columnwidth]{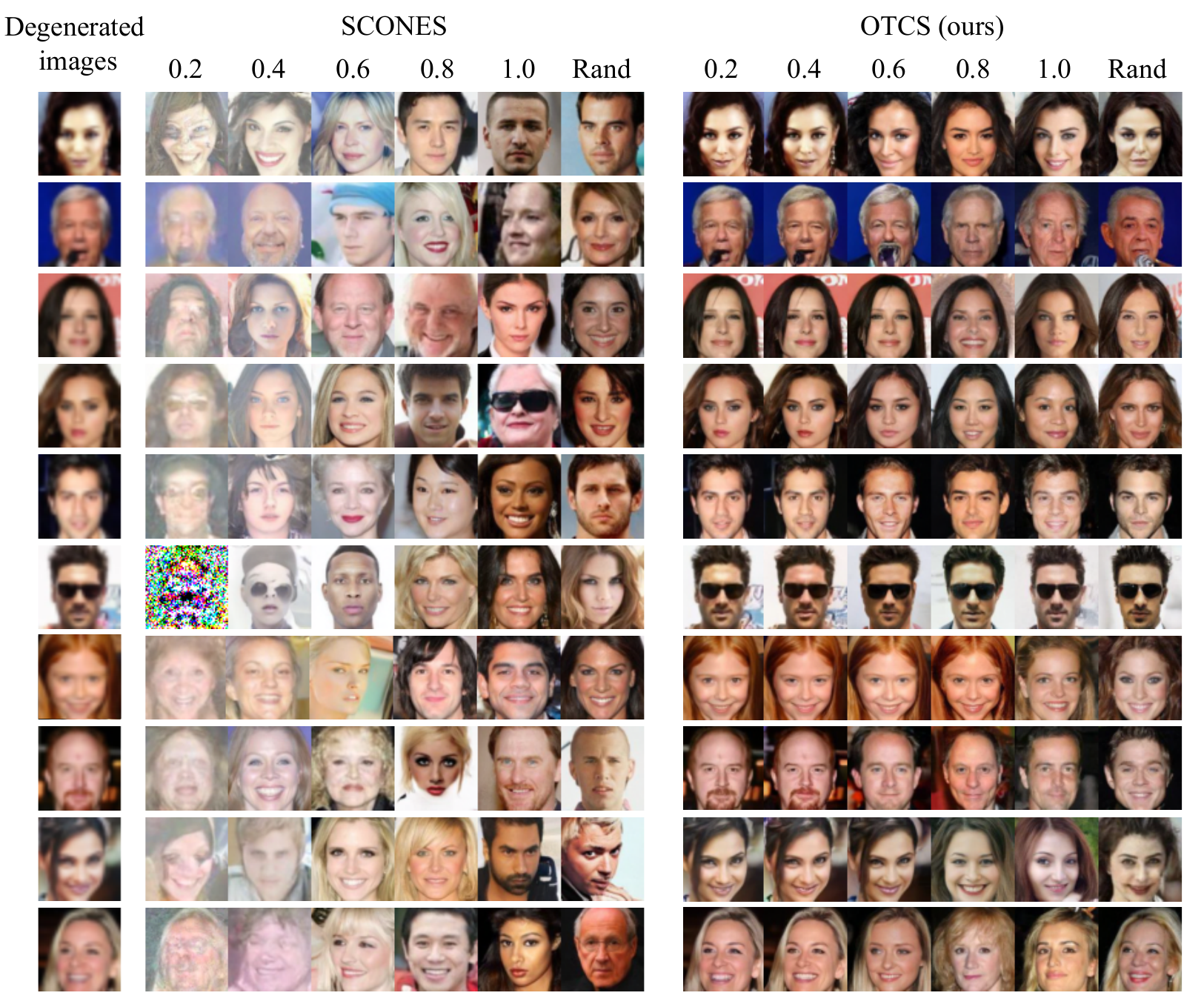} 
    \caption{Translated images by SCONES and OTCS with different initialization in inference. We consider the following initialization strategies: 1) We sample a noisy data $\y_{M}$ from $p_{M|0}(\y_M|\x)$ as initial state, and the reverse SDE starts at time $M$.  $\x$ is the degenerated image and $M$ is set to 0.2, 0.4, 0.8, and 1.0;  2) We directly generate a random noise $\y_{T}$ from $\mathcal{N}(0,\mathbf{I})$ as initial state (denoted as ``Rand'').
    }
    \label{fig_a:images_celeba}
\end{figure}
\paragraph{Translated images by SCONES and OTCS with different initial strategies in reverse SDE in inference.} We show the translated images of SCONES and our proposed OTCS in Fig.~\ref{fig_a:images_celeba} with different initialization strategies in inference. We can observe that for the smaller initial noise scales (0.2 and 0.4), the translated images by SCONES are not realistic. For larger initial noise scales (0.8 and 1.0),  the structures of translated images by SCONES are apparently different from those of degenerated images. The translated images by OTCS seem to be more realistic and share better structure similarity to degenerated images than SCONES, under different initial noise scales. 

\clearpage
\subsection{Additional Results in Semi-paired I2I}
    
\begin{figure}[H]
    \centering
    \includegraphics[width=1.0\columnwidth]{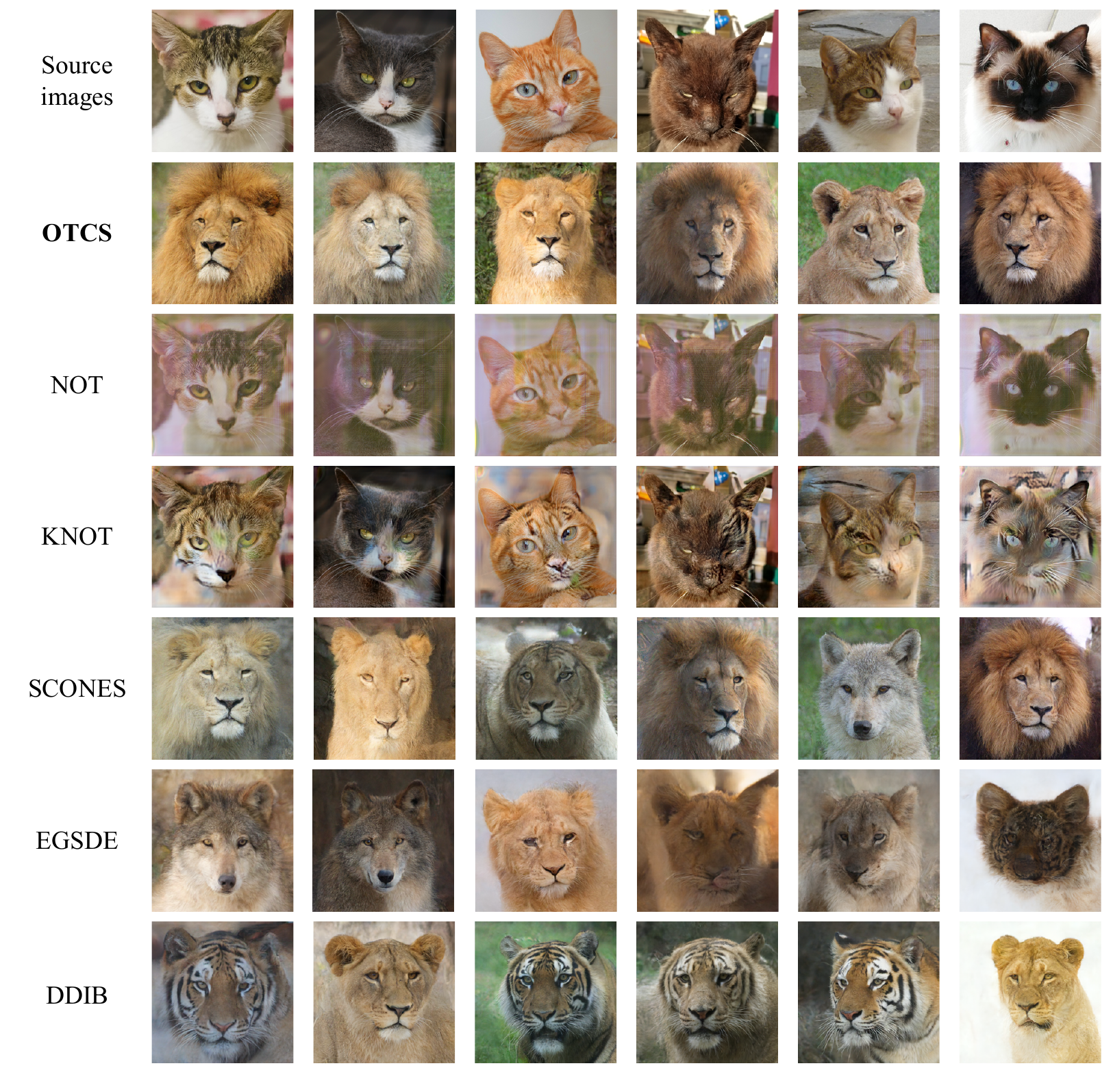} 
    \caption{Translated images of cat by different methods. With the guidance of paired images, we expect the images of cat to be translated into images of lion.
    }
    \label{fig_a:more_results_animal}
\end{figure}
\begin{figure}[H]
    \centering
    \includegraphics[width=1.0\columnwidth]{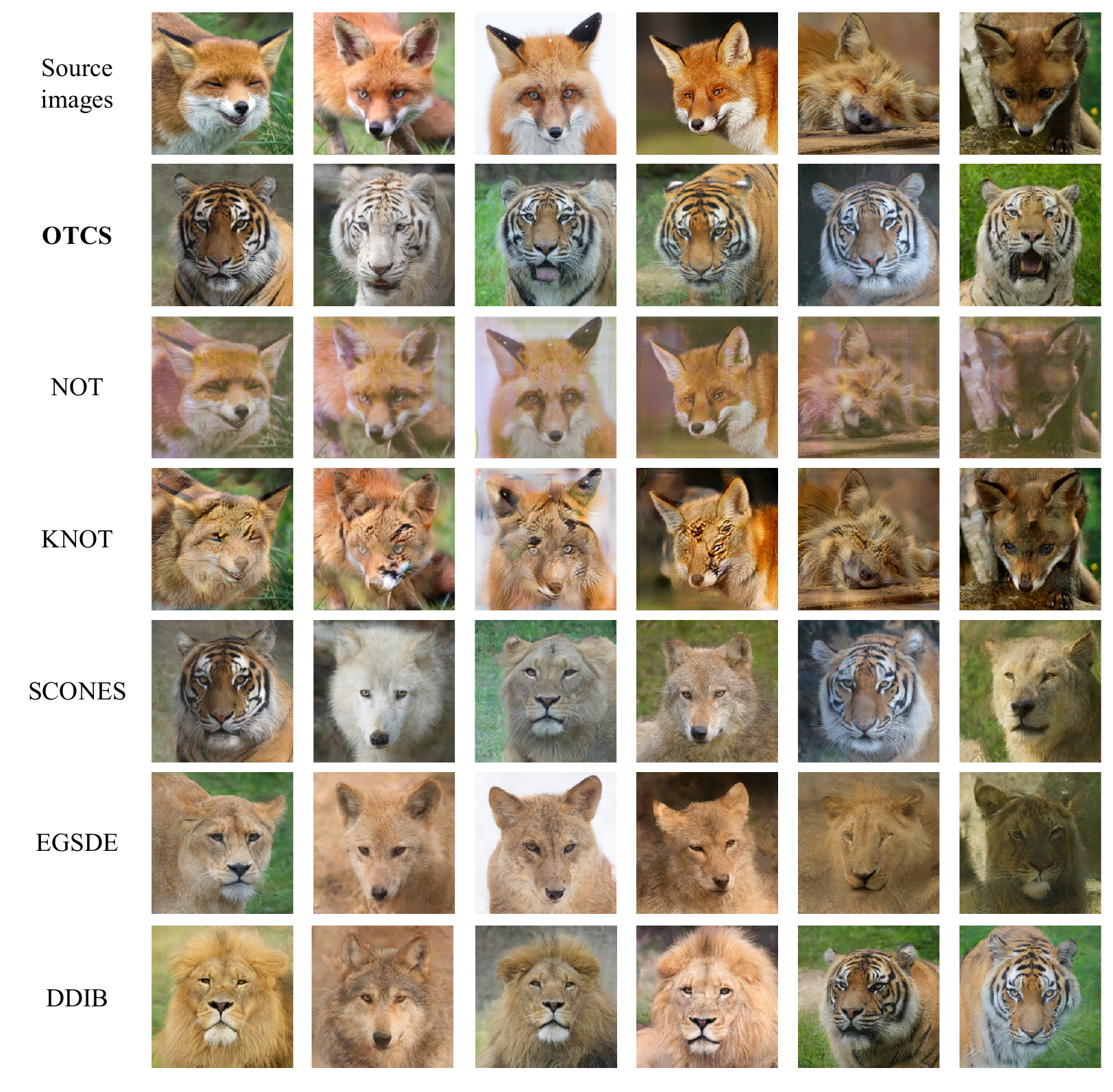} 
    \caption{Translated images of fox by different methods. With the guidance of paired images, we expect the images of fox to be translated into images of tiger.
    }
    \label{fig_a:more_results_animal1}
\end{figure}
\begin{figure}[H]
    \centering
    \includegraphics[width=1.0\columnwidth]{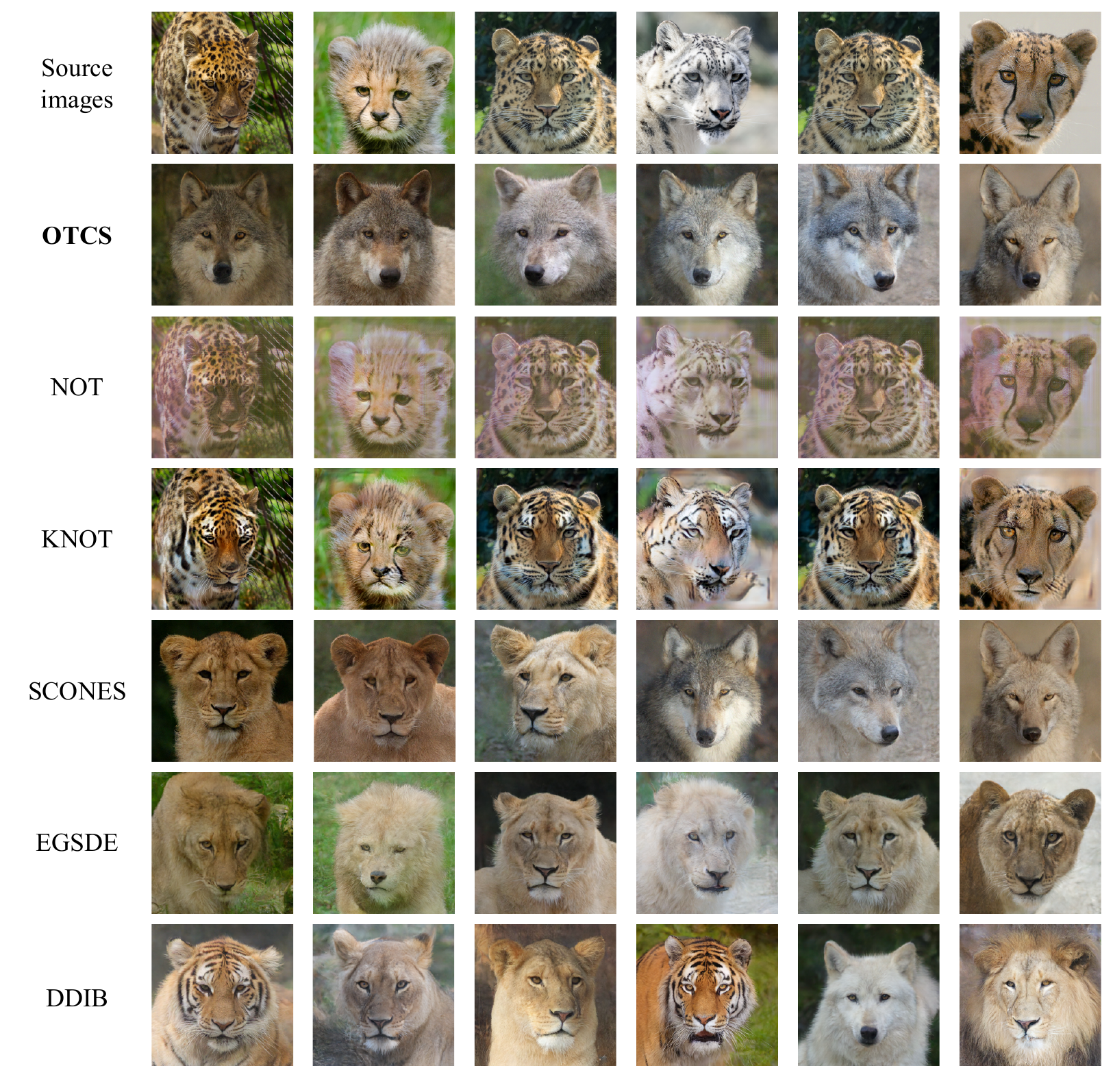} 
    \caption{Translated images of leopard by different methods. With the guidance of paired images, we expect the images of leopard to be translated into images of wolf.
    }
    \label{fig_a:more_results_animal2}
\end{figure}

\paragraph{Translated animal images by different methods.} We provide translated animal images by different methods in Figs.~\ref{fig_a:more_results_animal},~\ref{fig_a:more_results_animal1}, and~\ref{fig_a:more_results_animal2}. We can see that OTCS better translates the source images to high-quality target images of desired species than the other methods.

    

\clearpage
\subsection{Results for Trained Models at Different Training Steps}
\begin{figure}[t]
    \centering
    \includegraphics[width=1.0\columnwidth]{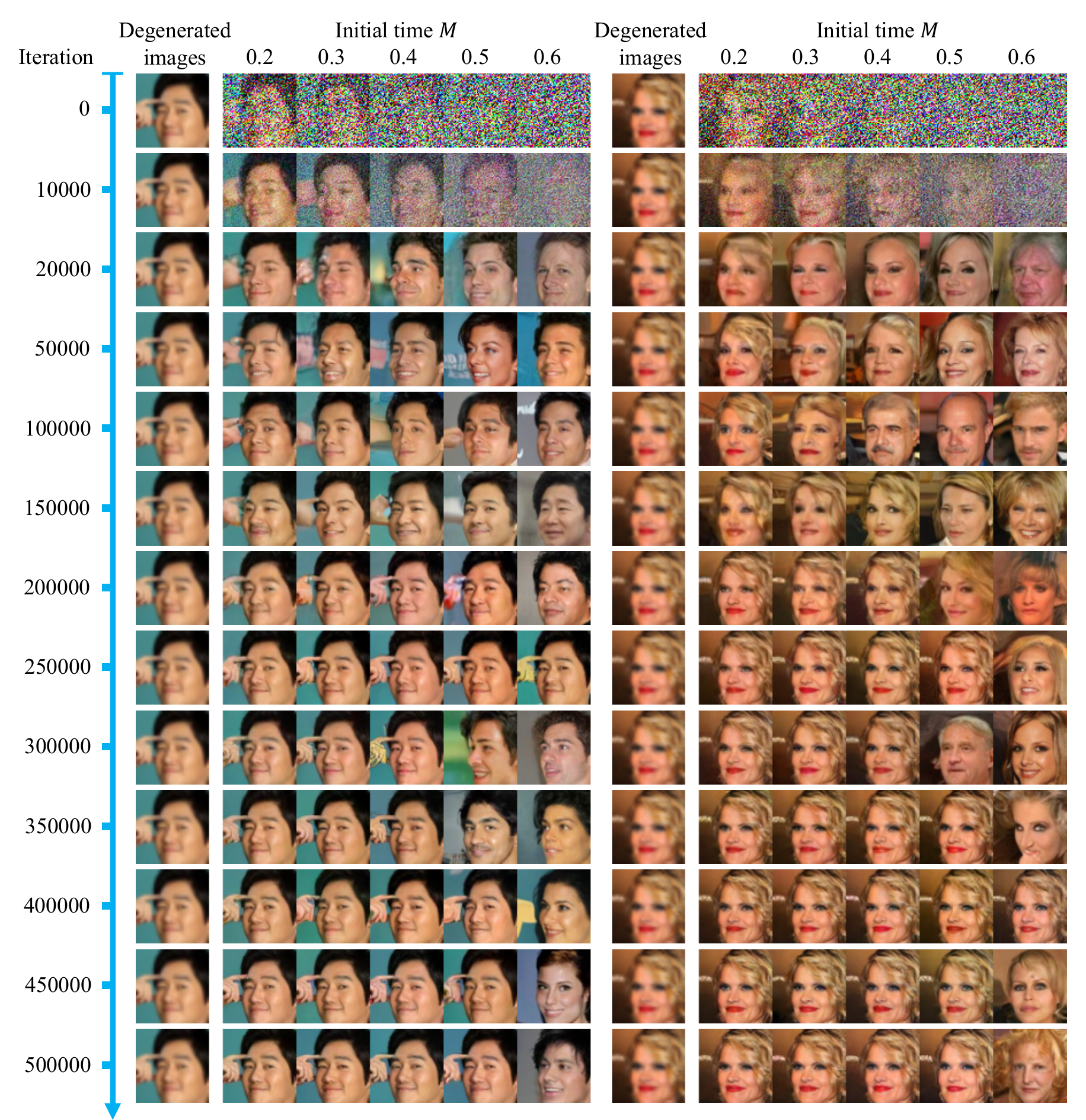}
    \caption{Translated images by OTCS using models at varying training steps, in which we consider different initial time in reverse SDE for generating samples.
    }
    \label{fig_a:images_different_steps}
\end{figure}

\begin{figure}[H]
    \centering
    \includegraphics[width=1.0\columnwidth]{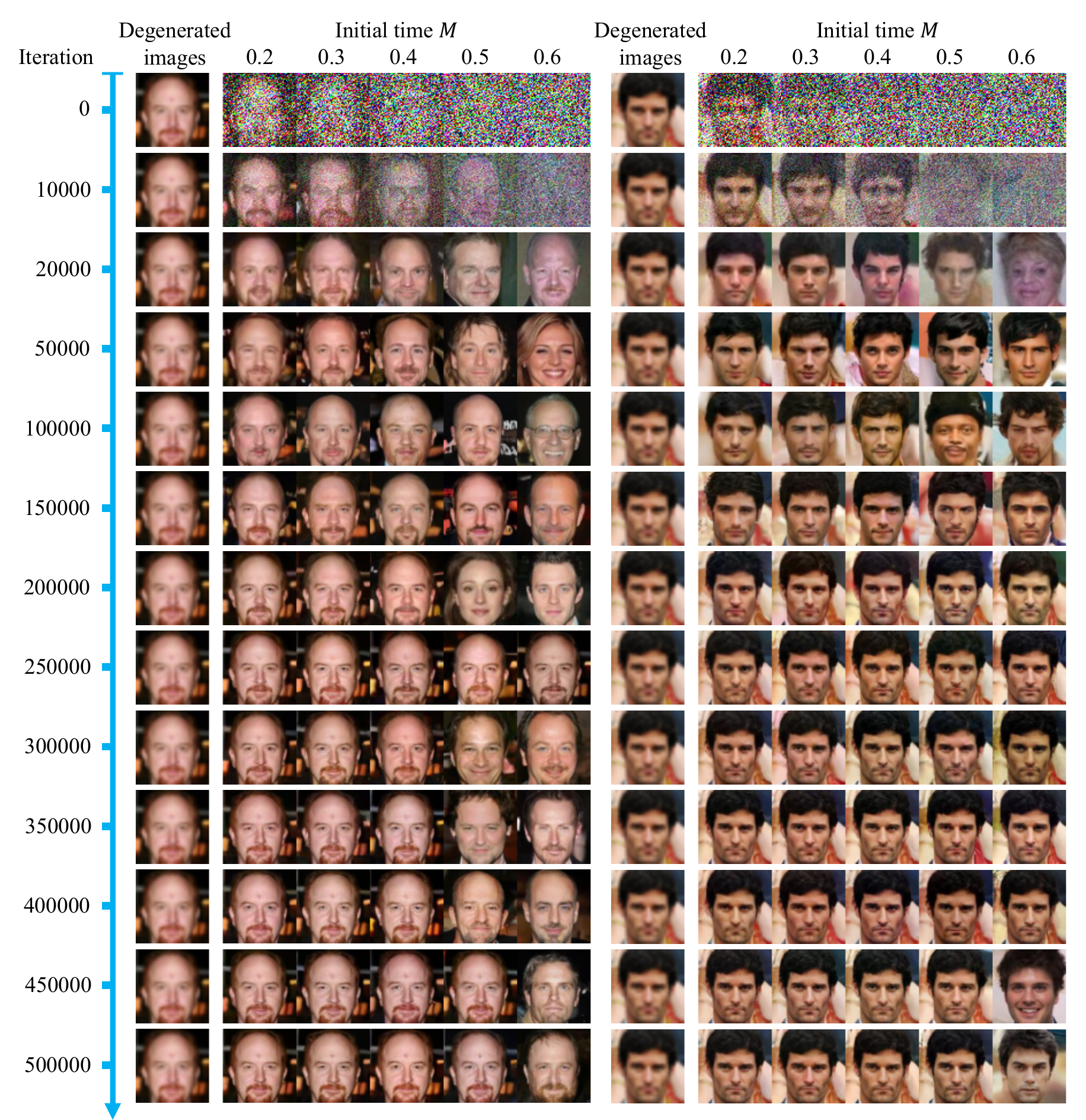}
    \caption{Translated images by OTCS using trained models at varying training steps, in which we consider different initial time in reverse SDE for generating samples.
    }
    \label{fig_a:images_different_steps1}
\end{figure}

\begin{figure}[H]
    \centering
    \includegraphics[width=1.0\columnwidth]{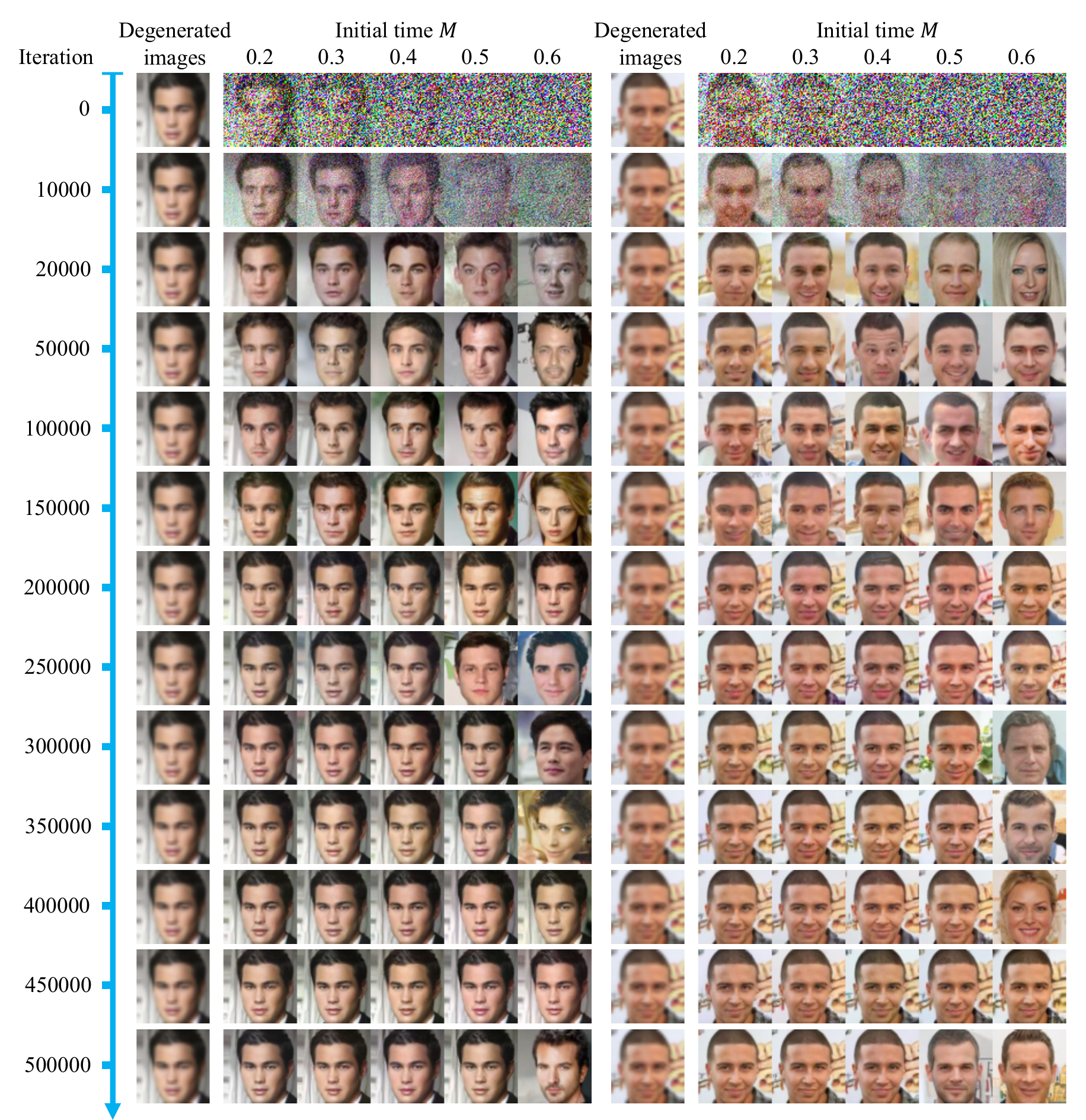}
    \caption{Translated images by OTCS using models at varying training steps, in which we consider different initial time in reverse SDE for generating samples.
    }
    \label{fig_a:images_different_steps2}
\end{figure}

In Figs.~\ref{fig_a:images_different_steps},~\ref{fig_a:images_different_steps1}, and~\ref{fig_a:images_different_steps2}, we show the translated images by OTCS in unpaired super-resolution using trained models at varying training steps, in which we consider different initial time $M$ in reverse SDE for generating samples.
\vspace{4cm}
\quad


\end{document}